\documentclass[runningheads]{llncs}
\usepackage{tikz}
\usepackage{comment}
\usepackage[T1]{fontenc}
\pdfoutput=1%
\usepackage{booktabs}
\newcommand{\matr}[1]{\mathbf{#1}}
\usepackage{algorithmic}
\usepackage{float}
\usepackage{multirow}
\usepackage{tikz}
\usepackage{comment}
\usepackage{amsmath,amssymb} 
\usepackage{color}
\usepackage[figuresright]{rotating}
\usepackage{graphicx}
\usepackage{ulem}
\usepackage{epsfig}
\usepackage{graphicx}
\usepackage{array}
\usepackage{makecell}
\usepackage{amsfonts}
\usepackage[ruled,lined]{algorithm2e}
\usepackage{cuted}
\usepackage{dblfloatfix}
\usepackage{epstopdf}
\usepackage{enumerate}
\usepackage{footnote}
\usepackage{gensymb}
\usepackage{multirow,booktabs}
\usepackage{longtable}
\usepackage{subfigure}
\usepackage{xcolor}
\usepackage{hyperref}
\hypersetup{
    colorlinks=true,
    urlcolor=magenta,
}
\usepackage[accsupp]{axessibility}  

\usepackage[width=122mm,left=12mm,paperwidth=146mm,height=193mm,top=12mm,paperheight=217mm]{geometry}

\newcommand{\deli}[1]{\textcolor{black}{#1}}
\newcommand{\xt}[1]{\textcolor{black}{#1}}
\newcommand{\xy}[1]{\textcolor{black}{#1}}
\let\emph\textit
\begin{document}

\pagestyle{headings}
\mainmatter
\def\ECCVSubNumber{4041}  

\title{UFO: Unified Feature Optimization} 

\titlerunning{ECCV-22 submission ID \ECCVSubNumber} 
\authorrunning{ECCV-22 submission ID \ECCVSubNumber} 
\author{Anonymous ECCV submission}
\institute{Paper ID \ECCVSubNumber}

\titlerunning{UFO: Unified Feature Optimization}

\author{Teng Xi\inst{}\thanks{Equal contribution} \and
Yifan Sun\inst{\star} \and
Deli Yu\inst{\star} \and Bi Li\inst{\star} \and Nan Peng\inst{\star} \and Gang Zhang\inst{}\thanks{Corresponding to \{zhanggang03,xiteng01\}@baidu.com}  \and \\ Xinyu Zhang \and Zhigang Wang \and Jinwen Chen \and Jian Wang \and Lufei Liu \and \\ Haocheng Feng  \and Junyu Han  \and Jingtuo Liu  \and Errui Ding  \and Jingdong Wang }

\authorrunning{Teng Xi, Yifan Sun, Deli Yu, Bi Li, Nan Peng, Gang Zhang et al.}

\institute{Baidu Inc. \\
Code: \color{magenta}{\url{https://github.com/PaddlePaddle/VIMER/tree/main/UFO}}}

\maketitle

\begin{abstract}

This paper proposes a novel Unified Feature Optimization (UFO) paradigm for training and deploying deep models under \xy{real-world and large-scale scenarios,} which requires a collection of multiple AI functions. UFO aims to benefit each \xy{single} task with a large-scale pretraining on all tasks. Compared with \xy{existing foundation models},
UFO has two points of emphasis, \emph{i.e.}, relatively smaller model size and NO adaptation cost: 1) UFO squeezes a wide range of tasks into a
moderate-sized unified model in a multi-task learning manner and further trims the model size when transferred to down-stream tasks. 2) UFO does not emphasize transfer to novel tasks. Instead, it aims to make the trimmed model dedicated for one or more already-seen task. To this end, it directly selects partial modules in the unified model, \xy{requiring}
completely NO adaptation cost. With these two characteristics, UFO provides great convenience for flexible deployment, while maintaining the benefits of large-scale pretraining. A key merit of UFO is that the trimming process not only reduces the model size and inference consumption, but also \xt{even} improves the accuracy on \xt{certain} tasks. Specifically, UFO considers the multi-task training \deli{and} brings \xy{a} two-fold impact on the unified model: some closely-related tasks have mutual benefits, while some tasks have conflicts against each other. UFO manages to reduce the conflicts and preserve the mutual benefits through a novel Network Architecture Search (NAS) method. 
Experiments on a wide range of deep representation learning tasks (\emph{i.e.}, face recognition, person re-identification, vehicle re-identification and product retrieval) show that the model trimmed from UFO achieves higher accuracy than its single-task-trained counterpart and yet has smaller model size, validating the concept of UFO. 
\xt{Besides, UFO also supported the release of \href{https://github.com/PaddlePaddle/VIMER/tree/main/UFO}{17 billion parameters computer vision (CV) foundation model} which is the largest CV model in the industry.}  

\end{abstract}

\section{Introduction}

Training and deploying are two essential procedures for artificial intelligence (AI) applications based on deep learning. A realistic AI system usually consists of multiple tasks. The naive train-and-deploy strategy is to train a respective deep model on each single sub-task for individual deployment. Given that some sub-tasks are actually correlated, this naive strategy wastes their mutual benefits.
A feasible approach to benefit individual tasks with the large-scale multi-task data is the foundation model. In this paper, we refer the foundation model as ``a model that is trained on broad data at scale and can be adapted to a wide range of downstream tasks'', according to \cite{bommasani2021opportunities}.

However, foundation model has some burden for deployment, \emph{e.g.}, it maintains the huge foundation model size and requires additional adaptation costs when transferred to down-stream tasks. 

This paper presents a novel train-and-deploy paradigm, named Unified Feature Optimization (UFO), to benefit down-stream tasks with large-scale multi-task pretraining. Compared to foundation model, UFO has two different points of emphasis, \emph{i.e.}, relatively smaller model size and NO adaptation cost. \textbf{1)} \emph{Small model size.} UFO does not use a tremendous network. Instead, it squeezes a wide range of tasks into a moderate-sized unified model, and further trims the model size for down-stream applications, so that the inference will be more efficient. \textbf{2)} \emph{No adaptation cost.} 
UFO does not emphasize transferring to novel tasks. Instead, it aims to make the trimmed model dedicated for already-seen sub-tasks. Without fine-tuning or prompt-based learning, UFO directly selects partial components from the already-learned unified model and thus requires completely no adaptation cost.

With the advantages of small model size and no adaptation cost, UFO provides great convenience for flexible deployment while maintaining the benefits of large-scale pretraining. Although the advantage of no adaptation cost is constrained to the already-seen sub-tasks, it does \deli{compromise} great benefits for realistic AI development. For example, in the smart city prototype, \textcolor{black}{like vision-based smart city, the system needs the collaboration of face, body and car to provide comprehensive understanding of the state of the city.}
Moreover, in spite that UFO lays no emphasis on the mode of transferring to novel down-stream tasks, it is compatible to this mode through existing foundation model techniques, which is not the major concern of this paper. Given their orthogonal advantages, we believe UFO and foundation model can well co-operate with each other to bring another wave of development.

\begin{figure}[t]
\centering
 \scalebox{0.25}{
\includegraphics{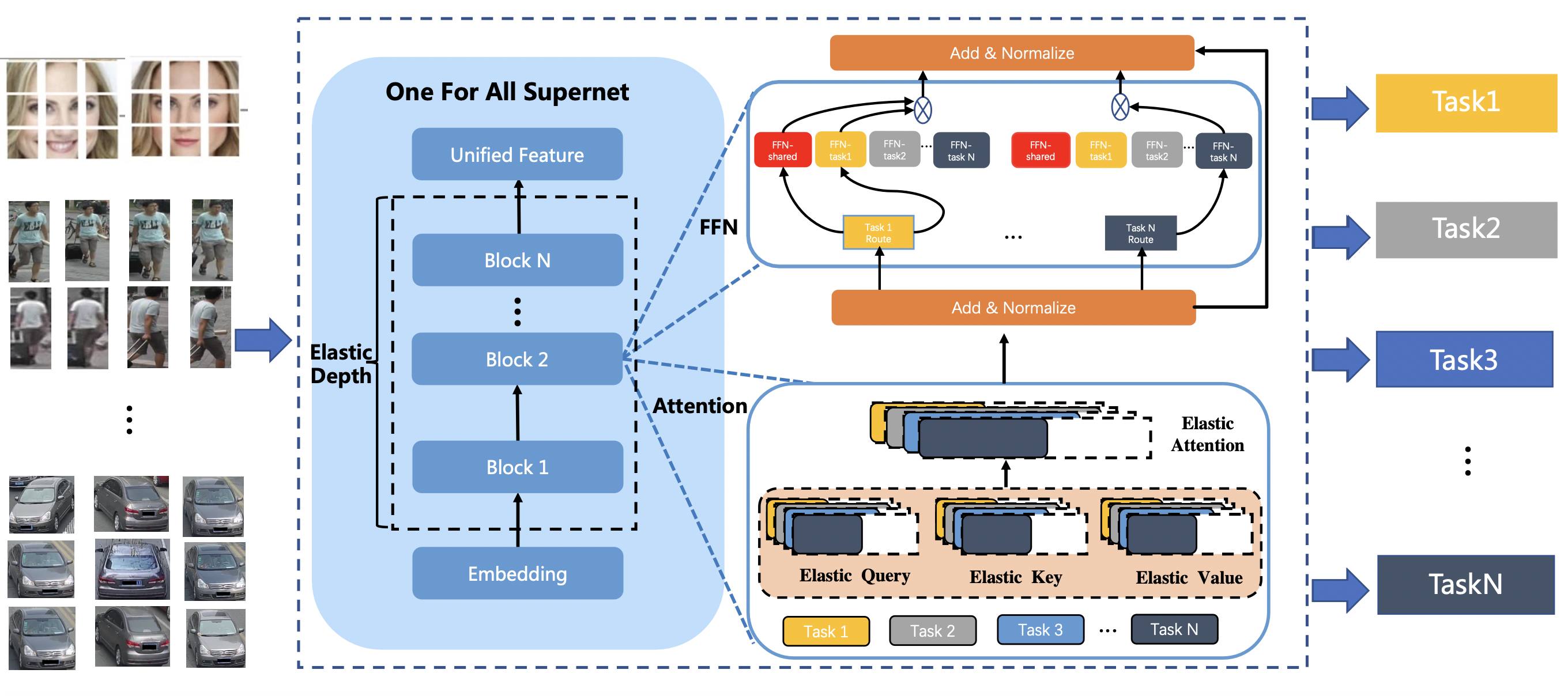}
}
\caption{Overview of the UFO paradigm. 
}
\label{fig:all}
\end{figure}

As an early exploration, this paper presents the concept of UFO with focus on deep representation learning, as shown in Fig. \ref{fig:all}. Deep representation learning is fundamental for a lot of AI applications, \emph{e.g.}, \textcolor{black}{face recognition \cite{an2021partial,kim2020groupface,chrysos2021deep}, person / vehicle re-identification \cite{herzog2021lightweight,herzog2021lightweight,he2021transreid,huynh2021strong,he2020fastreid} and fine-grained image retrieval \cite{lee2020compounding}. We base our UFO on the vision transformer (ViT) \cite{dosovitskiy2020image} architecture.} UFO first \deli{trains} a unified model (\emph{i.e.}, the supernet) on a variety of deep representation tasks in a multi-task learning manner. Afterwards, UFO learns to trim the supernet to get a dedicated sub-net 
for partial sub-tasks. Given a ViT backbone, \deli{the trimming object can be sub-block of the transformer, attention heads and FFN channels from coarse granularity to fine granularity, }
as illustrated in Fig.~\ref{fig:all}. \textcolor{black}{Moreover, UFO integrates another trimming strategy at the FFN path level. Following \cite{fedus2021switch}, UFO uses multiple FFN paths in parallel when training the supernet and allows trimming some FFN paths for down-stream tasks. Although these trimming strategies are popular, UFO is the first to integrate them and thus provides great trimming flexibility}. 

An important advantage of UFO is that the trimming process not only reduces the model size and inference consumption, but also improves the accuracy on its dedicated sub-tasks. It is non-trivial because trimming the model (without further fine-tuning) usually compromises the accuracy. To this end, UFO considers that the multi-task training brings two-fold \deli{impacts} on the supernet. On the one hand, some tasks are closely related to each other and thus have mutual benefits. On the other hand, some tasks have significant divergence and thus have mutual conflicts. During the trimming, UFO manages to reduce the conflicts and to preserve the mutual benefits through a novel Network Architecture Search (NAS) method. \textcolor{black}{Specifically, we design a search space for the UFO, which first \deli{introduces} FFN paths together with the supernet. \xy{Accordingly, we propose an end-to-end training strategy for UFO, which is different from previous multi-stage approaches \cite{cai2019once,hou2020dynabert}.}
Meanwhile, we also propose a novel evaluation metric for UFO, which is flexible to any requirements of practical application. } Experiments on a wide range of deep representation learning tasks show that UFO achieves higher accuracy with the smaller trimmed model than the single-task-trained counterpart. It confirms that while UFO gains the additional advantage of flexible deployment, it maintains the benefits of large-scale pretraining.

The contributions of the paper are summarized as follows:

\begin{itemize}

\item We propose a novel train-and-deploy paradigm, named Unified Feature Optimization (UFO), to benefit down-stream tasks with large-scale pretraining. UFO emphasizes the advantage of small model size and no adaptation cost, which significantly promotes flexible deployment. 

\item We propose a novel trimming process in UFO,  \deli{dedicated to} preserve the mutual benefits and discard the mutual conflicts from the multi-task unified model by \textcolor{black}{the proposed NAS method}.
 
 \item \textcolor{black}{We propose a novel evaluation metric to measure the correlations among tasks, which provides basic and effective \xy{analyses} for the trimming process.}
 
 \item We experiment on 10$+$ benchmarks from face, person, vehicle and product. Comprehensive \xy{analyses} and extensive experiments clearly show the effectiveness of our UFO.
 
\end{itemize}

\section{Related Work}

The development of smart city has led to significant demand \deli{on} the optimizations of multiple objectives to facility integrated solutions of diverse, real-world problems.  With an overall increase in number of models and tasks, significant computing and inference cost are required for deploying specific models for specific tasks, especially deployed on embedded sensors or devices where computational and power resources may be limited.  One way to solve this problem is the development of foundation models, which refer to models trained from broad data at scale that is capable of being adapted to a wide range of down-stream tasks. Existing works try to overcome these challenges from the following two aspects. 

\subsection{Training strategy}

Tuning weights of different task losses is an effective method. 
Kendall et al.~\cite{kendall2018multi} propose a principled approach to tune the weights of multiple loss functions by considering the homoscedastic uncertainty of each task. Dynamic Task Prioritization~\cite{guo2018dynamic} automatically prioritizes more difficult tasks by adaptively adjusting the mixing weight of each task’s loss objective. Other works adopt gradient-based methods to combat the challenge. GradNorm~\cite{chen2018gradnorm} automatically balances the training of different task losses in deep multi-task models by dynamically tuning their gradient magnitudes. 
Sener et al.~\cite{sener2018multi} explicitly cast multi-task learning as gradient-based multi-objective optimization, with the overall objective of finding a Pareto optimal solution to minimizing all task losses. Based on the observation that models with lower variance in the angles between task gradients  perform better, Suteu et al.~\cite{suteu2019regularizing} propose a novel gradient regularization of enforcing nearly orthogonal gradients. To avoid the interference of gradients from different losses, PCGrad~\cite{yu2020gradient} projects a task’s gradient onto the normal plane of the gradient of  other tasks that have a conﬂicting gradient. 

In contrast to these methods, our method designs a novel model structure, which adaptively specifies  correlations or conflicts among all tasks, and obtains competitive results even with ordinary training strategy.

\subsection{Model structure}

Some works~\cite{duong2015low,misra2016cross,liu2016recurrent,gao2020mtl} adopts the manner of soft parameter sharing. They allow each task to have separate model and parameters, but   enforce each model can access the information inside other models by regularizers~\cite{duong2015low,misra2016cross} or NAS-searched structures~\cite{gao2020mtl}. 

Other works~\cite{long2015learning,lu2017fully,subramanian2018learning,liu2019multi} use a shared part of backbone parameters with task-specific modules, which is called hard parameter sharing. 
 The first five convolutional layers are shared and task-specific fully-connected layers are used for each task in the method of Deep Relationship Networks\cite{long2015learning}. 
 Lu et al.~\cite{lu2017fully} starts with a thin network and dynamically grows it during the training phase by creating new branches for tasks.  
 Besides the area of computer vision, ~\cite{subramanian2018learning,liu2019multi} use shared encoders with task-specific layers across multiple NLP tasks. 

Beyond the two kind of ways, Task-MOE~\cite{kudugunta2021beyond} proposes an architecture which combines both the shared and task-specific modules for multi-task learning. Specifically, it shares the Self-Attention modules and selects task-specific FFN modules based on a task-level router. 

All these works consider adding components by encouraging the information interaction between single  tasks or introducing task-specific modules, but miss the idea of reducing modules. By contrast, we extract subnet by reducing incompatible weights and keeping complementary weights from a surpernet. 
Similar to Task-MOE, our method also adopts task-level routers to select specific FFNs. However, our method  extracts the most suitable sub-weights of Self-Attention for each task, while Task-MOE share the complete one among all tasks. 

\section{Methodology}

UFO consists of two steps, \emph{i.e.}, training a multi-task supernet, and extracting a dedicated sub-network for down-stream task deployment. Under this novel training and deploying paradigm, 
UFO aims to preserve the mutual benefit of multi-task pretraining and remove the mutual conflict between different tasks. To this end, we employ a Neural Architecture Search (NAS) method to search for the sub-network from the supernet. Specifically, we introduce the architecture of UFO supernet as well as its search space in Section \textcolor{red}{\ref{sec:3.1}}. We note that different from \xt{the search space} for single task, \xt{the UFO search space} is to accommodate multiple sub-networks for various downstream tasks. Given the architecture of UFO supernet, Section \textcolor{red}{\ref{sec:3.2}} explains how to train the supernet on all the tasks in a multi-task learning manner. Finally, Section \textcolor{red}{\ref{sec:3.3}} elaborates on learning the sub-network extraction based on NAS. It allows UFO to directly extract a corresponding sub-network through architecture prediction, given the desired down-stream tasks (as well as the model size and inference speed). 

\subsection{The architecture and search space of UFO supernet}\label{sec:3.1}

As shown in figure \ref{fig:all} , we base the UFO supernet on the vision transformer (ViT). Since the sub-network selects partial modules from the supernet and inherits the corresponding parameters \deli{during the deployment}, it is important that the supernet \deli{provides} a large space for searching and extracting the sub-networks. 

Prior transformer-based NAS usually provides three searching directions, \emph{i.e.}, elastic depths, elastic attention heads and elastic expansion ratios of the Feed Forward Networks (FFN) \cite{kudugunta2021beyond}. In addition to these commonly-used searching directions, we introduce a novel search direction, \emph{i.e.}, flexible FFN paths. In other words, UFO combines three commonly-used search directions and a novel one, \deli{and thus provides} a large searching space. \deli{Consequently}, the sub networks can reduce FFN paths, FFN weights, attention weights or even the whole sub blocks of the vision transformer. We explain these searching directions in details as below.

The architecture space is consisted with a set of architectures, \deli{ and is} denoted as $\mathcal  A=\{a_1, a_2, \cdots, a_{n_a}\}$, $n_a$=$|\mathcal A|$.  Let $\mathcal H$ be set of head numbers and $\mathcal M$ be the set of mlp ratios in FFN, where $\mathcal H = \{h_1,h_2,\cdots,h_{n_h}\}$, $n_h$=$|\mathcal H|$ and $\mathcal M = \{m_1,m_2,\cdots,m_{n_m}\}$, $n_m$=$|\mathcal M|$. Let $\mathcal T$ be the set of target tasks, where $\mathcal T = \{t_1,t_2,\cdots,t_{n_t}\}$, $n_t$=$|\mathcal T|$.  Let $\mathcal G$ be the set of gate choice of FFN paths, where $\mathcal G = \{g_0,g_1,\cdots,g_{n_g}\}$, $n_g$=$|\mathcal G|$. Finally, let $\mathcal D$=\{0,1\} be the set of drop choice to denote whether the entire layer will be dropped. Then, the search space $\mathcal  A$ can be denoted as follows: $\mathcal A$ = $\{[ [h_1,m_1,g_1,d_1],[h_2,m_2,g_2,d_2],\cdots,[h_l,m_l,g_l,d_l]],h_i \in \mathcal H, m_i \in \mathcal M, g_i \xt{\subseteq} \mathcal G, d_i \in \mathcal D, \forall i \in \{1,2,\cdots,l\}\}$, where $l$ is the numbers of layers. In summary, $\mathcal G$ determines the FFN paths of different tasks of $\mathcal T$. Furthermore, $\mathcal H$ and $\mathcal M$ determine the model size of different sub networks. Besides, $\mathcal D$ controls the depth of sub networks to further reduce the model size. 

\deli{Given the input $\textbf{x}_{i}^t$ of task $t$, an arch $a$ is sampled from $\mathcal{A}$, and then the consecutive blocks of
the arch are computed as: } 
\begin{equation}
      \begin{array}{l} 
    \hat{\textbf{x}}_{i}^t = d_l * \text{MHSA}(\text{LN}(\textbf{x}_{i}^t), h_l)+  \textbf{x}_{i}^t\\ 
    \textbf{x}_{i+1}^t = d_l * \text{FFNs}(\text{LN}(\hat{\textbf{x}}_{i}^t), m_i, g_i^t) + \hat{\textbf{x}}_{i}^t  \\
      \end{array}
\end{equation}

\subsection{Multi-task training of the UFO supernet }\label{sec:3.2}

In this subsection, we will describe how to train \deli{multi-task} supernet. As shown in subsection \ref{sec:3.1}, the supernet in UFO is quite different from other \deli{single-task supernets}. Accordingly, the training strategy of UFO is also different in two aspects of sub-network sampling and data sampling.

\subsubsection{Sub-network sampling.}
\deli{The sub-network sampling is involved with the sampling of $(m_l, h_l, d_l, g_l)$. Similar to weight entanglement mechanism \cite{chen2021autoformer}, the weights of the arch $a$ are shared with the weight of supernet for their common parts with respect to the sampling of $m_l$ and $g_l$.}
\deli{However, as the supernet do not have FFN-paths in the existing training strategies \cite{chen2021autoformer,hou2020dynabert,wang2020hat},} there are serious competitions among shared attention weights. Thus, \deli{their} supernet has to be trained in a step-by-step way. In the UFO, the FFN-paths relieve the competition of shared attention. Thus, the UFO can be trained in an end-to-end way. 

However, the supernet is hard to converge if we directly sample sub-networks from $\mathcal A$ with respect to $g_l$, because the total number of FFN paths is $|\mathcal T|\times(2^{|\mathcal{G}|}-1)^l$. Thus, we set constraint for the path gate of each task, where each task in $\mathcal T$ only \deli{has} 3 \deli{choices} for each layer, \deli{i.e.} shared FFN only, task specific FFN only or both. \xt{To be specific, we use gumbel-softmax on learn-able gate weights  to sample probability distribution for task $t$. Thus, the output of FFNs (we ignore layer/block idx $i$) can be defined as:}

\begin{equation}
      \begin{array}{l} 
\text{FFNs}(\cdot) =  p^t[0] \text{FFN}_{shared}(\cdot) + p^t[1] \xt{\text{FFN}_{task-specific}^t(\cdot)} \\
      \end{array}
\end{equation}

\xt{After training, the learned gate weights determine the single choice of gates by $\text{argmax}$ or choose both gates.} In this way, the total number of FFN paths is reduced from $|\mathcal T|\times(2^{|\mathcal{G}|}-1)^l$ to $|\mathcal T|\times|3|^l$.

\subsubsection{Data sampling}

\deli{There are five existing data sampling strategies in \cite{abnar2021exploring}. The accumulating gradient strategy is the most promising among them. It accumulates gradients from all task data in one optimizer step, and can achieve better optimization trade-off between different tasks than other methods, e.g. task-by-task and alternating methods.
Inspire by the thought, we propose a similar but different strategy of forming batch, and it is called heterogeneous batch type. 
}
To be specific,  we sample some data from all \deli{tasks} of $\mathcal T$ to form a mini batch with a weight roughly proportional to the size of the task datasets, respectively. 
Then, these mini batch is concatenated into a batch data, which is feed into the backbone. Next, the obtained features are separated and feed into $|\mathcal T|$ task-specific head networks, each of which is responsible for
the output of a task. Finally, we calculate the loss of $|\mathcal T|$ tasks, sum it up for the shared transform backbone network, and finish a backward step to obtain gradients, which are used to update the shared parameters. 

\subsection{Extracting the sub-network for down-stream task deploying}\label{sec:3.3}
\xt{In this subsection, we will introduce how to select optimal dedicated models from supernet according to requirement of practical applications.}

Our target is to find optimal architecture $a$ of $\mathcal A$ under flops and parameter constraints and the average performance is maximized.

Let $f_{{t}}$($a$) be the performance of architecture $a$ on task  $t$, $\forall t \in \mathcal T$, $\forall a \in \mathcal A$. Then, let $f_{\matr{t}}$($a$) be the performance of architecture $a$ on task set  $\matr t$, $\forall \matr{t} \subset \mathcal T$, $\forall a \in \mathcal A$.

Except for the extreme performances for target tasks, we also care about the generalized performances on other tasks. Thus, let $avg\_f(\matr{t}, a)$ be the comprehensive performance of architecture on all tasks, where:
\begin{align}
avg\_f(\matr{t}, a) &= \lambda f_{\matr t}(a) + (1 - \lambda) f_{\mathcal T \setminus \matr t}(a) \\
& = \lambda \sum_{t_1 \in \matr t}f_{t_1}(a) + (1-\lambda) \sum_{t_2\in \mathcal T \setminus \matr t}f_{t_2}(a) \\
& ,\forall \matr{t} \subset \mathcal T, \forall a \in \mathcal A
\end{align}

\xt{$\lambda$ is set to $1/|\mathcal T|$ by default, and can be flexibly adjusted according to different tasks.}

Then, we can formulate as follows: 
\begin{align}\label{eq:pair_wise_formulation2_new}
\text{max} \ \ & 
\lambda \sum_{t_1 \in \matr t}f_{t_1}(a) + (1-\lambda) \sum_{t_2\in \mathcal T \setminus \matr t}f_{t_2}(a)\\
\text{s.t.} \ \ & 
 a \in \mathcal A,  \\
 & flops(a) <= constraint\_flops \\
 & parameters(a) <= constraint\_parameters
\end{align}

\begin{algorithm}[t] \label{MSA}
\footnotesize
\caption{Multi-task Searching Algorithm (MSA)}
\label{alg:MSA}
\DontPrintSemicolon
\LinesNumbered

\KwIn{
  $\mathcal T$, $\mathcal A$, $\matr t$;

}
\KwOut{
$a\_best$;}

  Generate a sub set of architectures $\mathcal S$ from $\mathcal A$, $\mathcal S \subset \mathcal A$.  \\
  Initialize performance predictors $pre(a,t)$ for each task, $\forall t \in \mathcal T$, $\forall a \in \mathcal A$. \\
  Initialize $ready\_flag\_t$  for each task, $\forall t \in \mathcal T$. \\
  \For{$n=1;n<=k;n++$}
  {
            Sample architectures from $\mathcal S$. \\ 
            \For{$t \in \mathcal T$}
            {
                   Train predictor of task $t$. \\
                   Calculate predicted architectures ranks of task $t$. \\
                   Calculate Kendall tau between ground truth ranks and predicted ranks for task $t$, $kd\_t$. \\
                   \If{$kd\_t >=$ thre\_t}
                   {
                     $ready\_flag\_t$ = 1; \\
                   }
            }
          \If{$ready\_flag\_t == 1, \forall t \in \mathcal T \ or \ n==k$}
               {
                 Calculate objective function of ORP for all architectures in $\mathcal S$ according to Eq.~\ref{obj_rank}. \\
                 Select best architecture $a\_best$. \\
                 Return $a\_best$. \\
               }
          
  }
 
\end{algorithm}

\xt{Nevertheless, as} $avg\_f(\matr{t}, a)$ has different metrics for different \deli{tasks} which can not be added directly, \xt{we use rank instead of performance.}

Similarly, let $r_t$($a$) be the rank of performance of architecture $a$ on task $t$, $\forall t \in \mathcal T$, $\forall a \in \mathcal A$. Then, let $r_{\matr{t}}$($a$) be the rank of performance of architecture $a$ on task set $\matr {t}$, $\forall \matr{t} \subset \mathcal T$, $\forall a \in \mathcal A$.

\xt{Similarly}, we also care about the generalized rank on other tasks. 
\xt{Accordingly}, let $avg\_r(\matr{t}, a)$ be the comprehensive rank of architecture on all tasks, where:
\begin{align}
avg\_r(\matr{t}, a) &= \lambda r_{\matr t}(a) + (1 - \lambda) r_{\mathcal T \setminus \matr t}(a) \\
& =  \lambda \sum_{t_1 \in \matr t}r_{t_1}(a) + (1-\lambda) \sum_{t_2\in \mathcal T \setminus \matr t}r_{t_2}(a) \\
& ,\forall \matr{t} \subset \mathcal T, \forall a \in \mathcal A
\end{align}

\xt{Finally}, we formulate the optimal rank problem (ORP) as follows: 
\begin{align}\label{obj_rank}
\text{min} \ \ 
&  \lambda \sum_{t_1 \in \matr t}r_{t_1}(a) + (1-\lambda) \sum_{t_2\in \mathcal T \setminus \matr t}r_{t_2}(a)   \\ 
\text{s.t.} \ \ 
 & a \in \mathcal A,  \\
 & flops(a) <= constraint\_flops \\
 & parameters(a) <= constraint\_parameters
\end{align} 

Then, a multi-task searching algorithm (MSA) is proposed to solve the ORP. Alogrithm \ref{MSA} shows the pseudo code of MSA. As the search space is huge, we first generate a sub set of architectures $\mathcal S$ from $\mathcal A$, $\mathcal S \in \mathcal A$. Then, we sample architectures from $\mathcal S$  to train task specific performance predictors separately \xt{based on GP-NAS \cite{li2020gp}}. We utilize Kendall tau to measure the accuracy of the predictors. When the predictors are all well trained, we calculate the objective function of ORP for all architectures in $\mathcal S$ according to Eq.~\ref{obj_rank} and select best architecture $a\_best$. In the experiment section, we will evaluate the performance of  task specific predictors thoroughly.  

\section{Experiments}

\subsection{Settings}

\begin{table*}[b!]
\caption{Trainning Dataset}
\vspace{-4mm}
\label{table:Tranning}
\begin{center}
\begin{tabular}{cccccc}
\hline
Tasks &Datesets & Img Number & ID Number \\
\hline
Face &  MS1M-V3   & 5,179,510  & 93,431   \\
\hline
Person & Market1501-Train &12,936  & 751  \\
\hline
Person & MSMT17-Train  &  30,248 & 1,041  \\
\hline
Vehicle & Veri-776-Train  &  37,778    & 576  \\
\hline
Vehicle & VehicleID-Train  &113,346  &13,164  \\
\hline
Vehicle & VeriWild-Train  & 277,797 & 30,671 \\
\hline
Products  & SOP-Train  & 59,551  & 11,318  \\

\hline
\end{tabular}
\end{center}
\vspace{-3mm}
\end{table*}

\begin{table*}[t]
\caption{Test Dataset}
\vspace{-4mm}
\label{table:Test}
\begin{center}
\begin{tabular}{cccc}
\hline
Tasks &Datesets & Img Number & ID Number \\
\hline
Face &  LFW   & 12,000   & -   \\
\hline
Face &  CPLFW   & 12,000   & -   \\
\hline
Face &  CFP-FF     & 14,000   & -   \\
\hline
Face &  CFP-FP   & 14,000   & -   \\
\hline
Face &  CALFW    & 12,000   & -   \\
\hline
Face &  AGEDB-30    & 12,000   & -   \\
\hline
Person & Market1501-Test &19,281  & 750   \\
\hline
Person & MSMT17-Test  &   93,820  & 3,060  \\
\hline
Vehicle & Veri-776-Test  &  13,257    & 200   \\
\hline
Vehicle & VehicleID-Test  &19,777  &2,400   \\
\hline
Vehicle & VeriWild-Test  & 138,517 &  10,000  \\
\hline
Products  & SOP-Test  & 60,502   & 11,316   \\

\hline
\end{tabular}
\end{center}
\vspace{-3mm}
\end{table*}

\begin{table*}[h]
\caption{Training Configurations}
\vspace{-4mm}
\label{table:setting_config}
\begin{center}
\begin{tabular}{cc}
\hline
 &Face/Person/Vehicle/Products \\
\hline
Input Size &  256 $\times$ 256   \\
\hline
Batch Size & 1024/512/512/512 \\
\hline
Augmentation & Flipping + Random Erasing + AutoAug  \\
\hline
Model & ViT-base   \\
\hline
Feature Dim & 768 \\
\hline
Loss & CosFace Loss/(CosFace Loss + Triplet Loss)*3  \\
\hline
Optimizer & SGD \\
\hline
Init LR  & 0.2  \\
\hline
LR scheduler  & Warmup + Cosine LR  \\
\hline
Iterations & 100,000  \\

\hline
\end{tabular}

\end{center}
\vspace{-3mm}
\end{table*}

\begin{figure*}[h!] 
	\subfigure[Person benchmarks]{
		\begin{minipage}[t]{0.48\linewidth}
			\centering
			\includegraphics[width=0.99\linewidth]{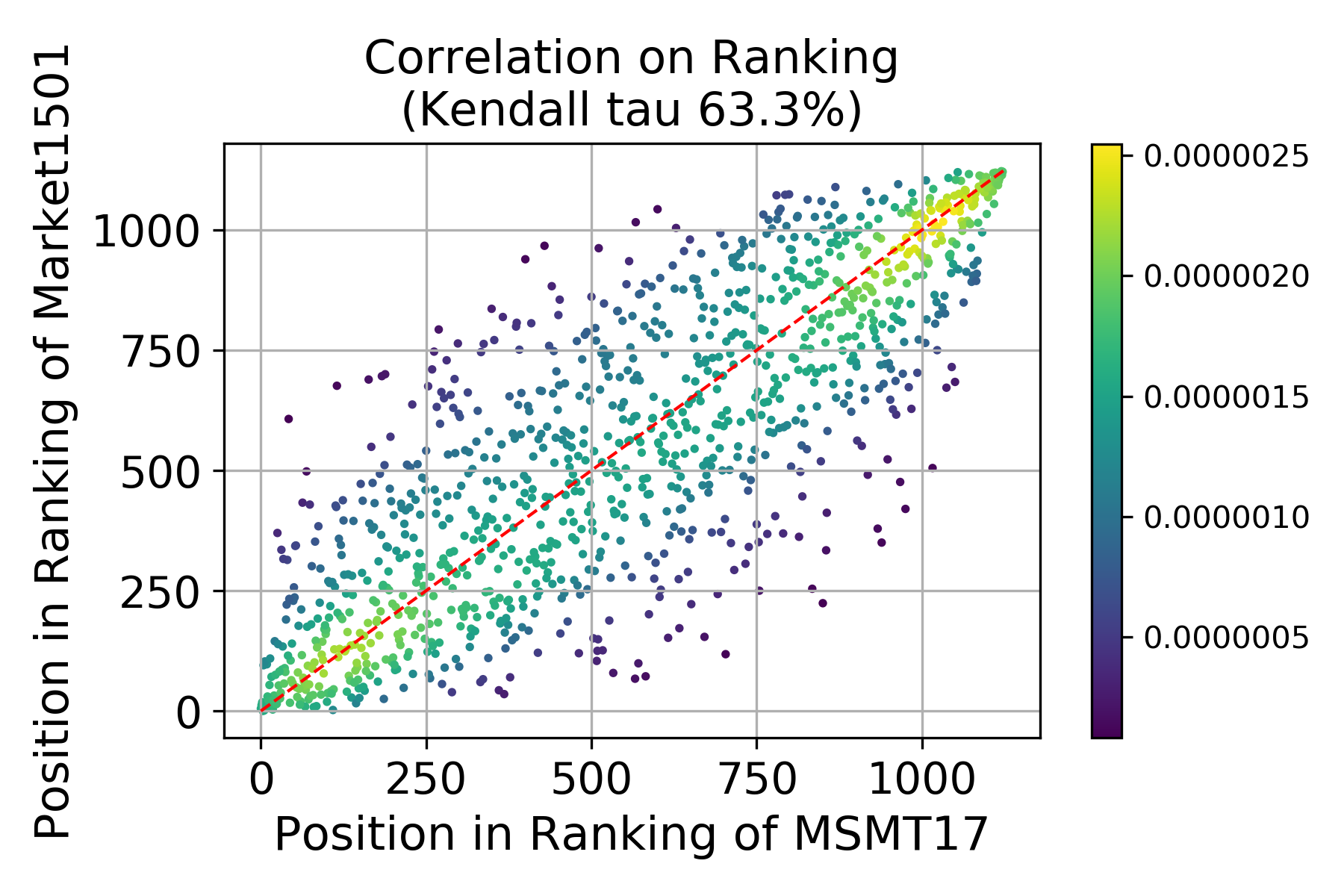}
		\end{minipage}
	}
	\subfigure[Vehicle benchmarks]{
		\begin{minipage}[t]{0.48\linewidth}
			\centering
			\includegraphics[width=0.99\linewidth]{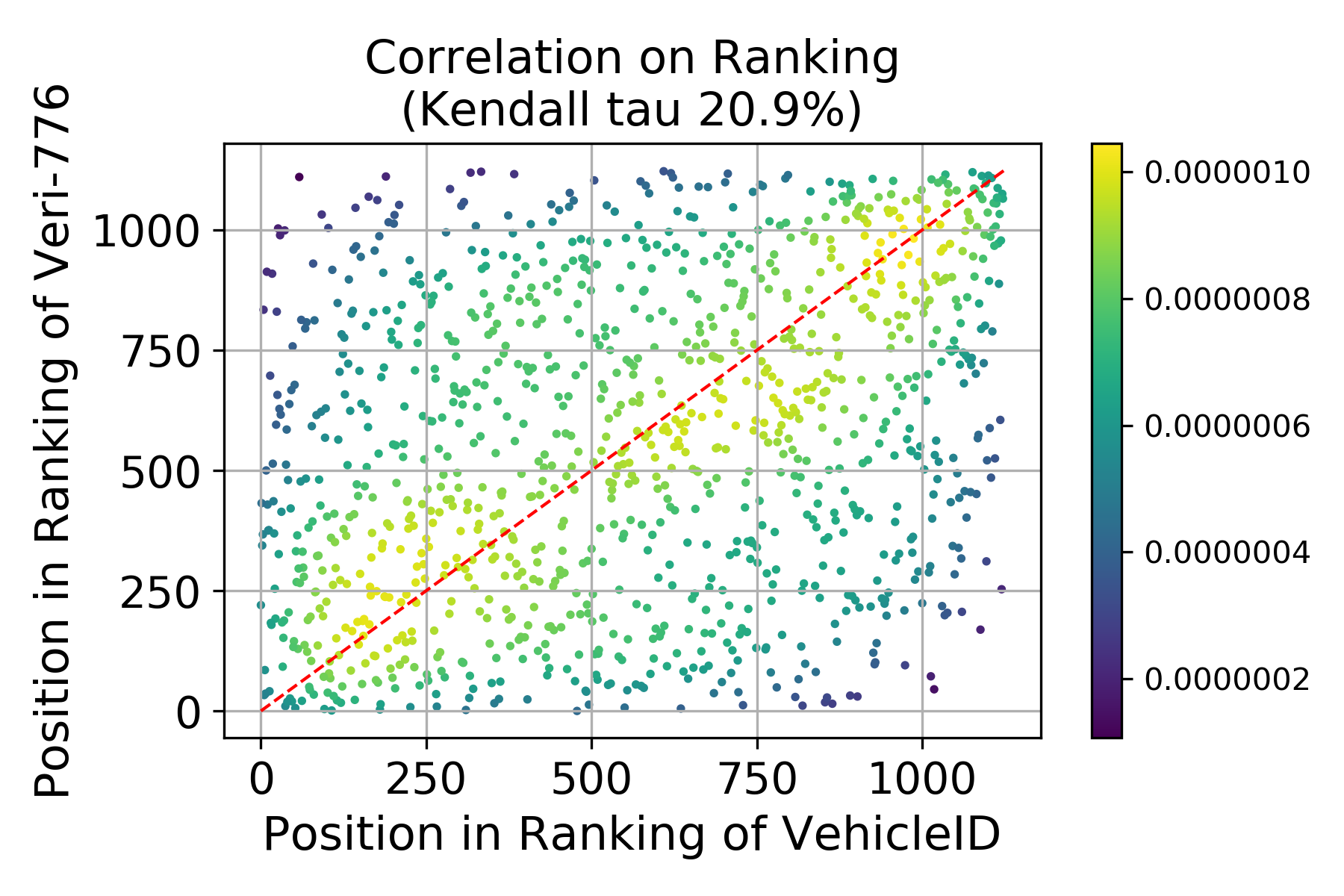}
		\end{minipage}
	} \ 
	\caption{Ranking correlation within tasks.}
	\label{within}
	\vspace{-8mm}
\end{figure*}

\begin{figure*}[h!] 
	\subfigure[Face and Person ]{
		\begin{minipage}[t]{0.48\linewidth}
			\centering
			\includegraphics[width=0.99\linewidth]{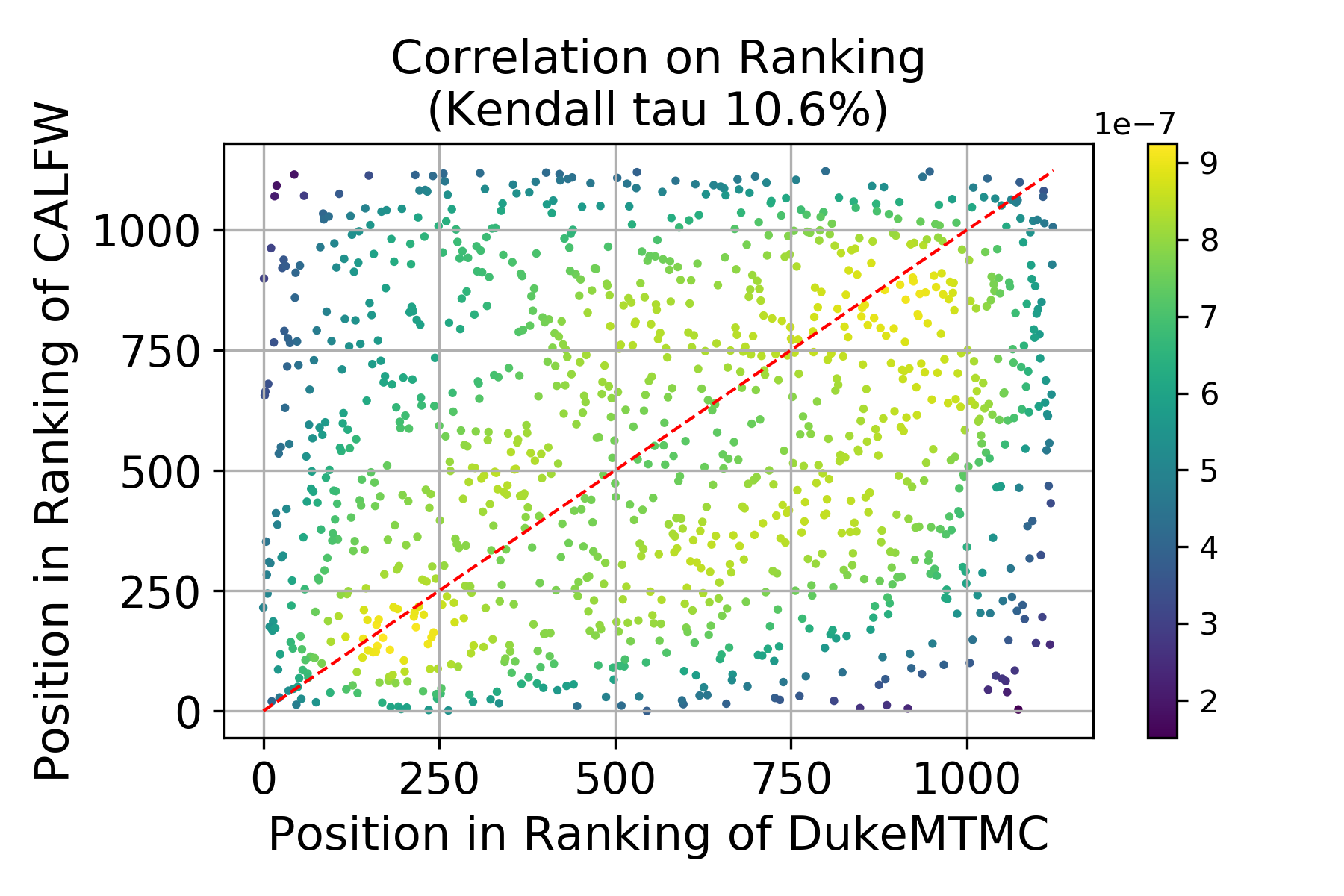}
		\end{minipage}
	}
	\subfigure[Face and Vehicle]{
		\begin{minipage}[t]{0.48\linewidth}
			\centering
			\includegraphics[width=0.99\linewidth]{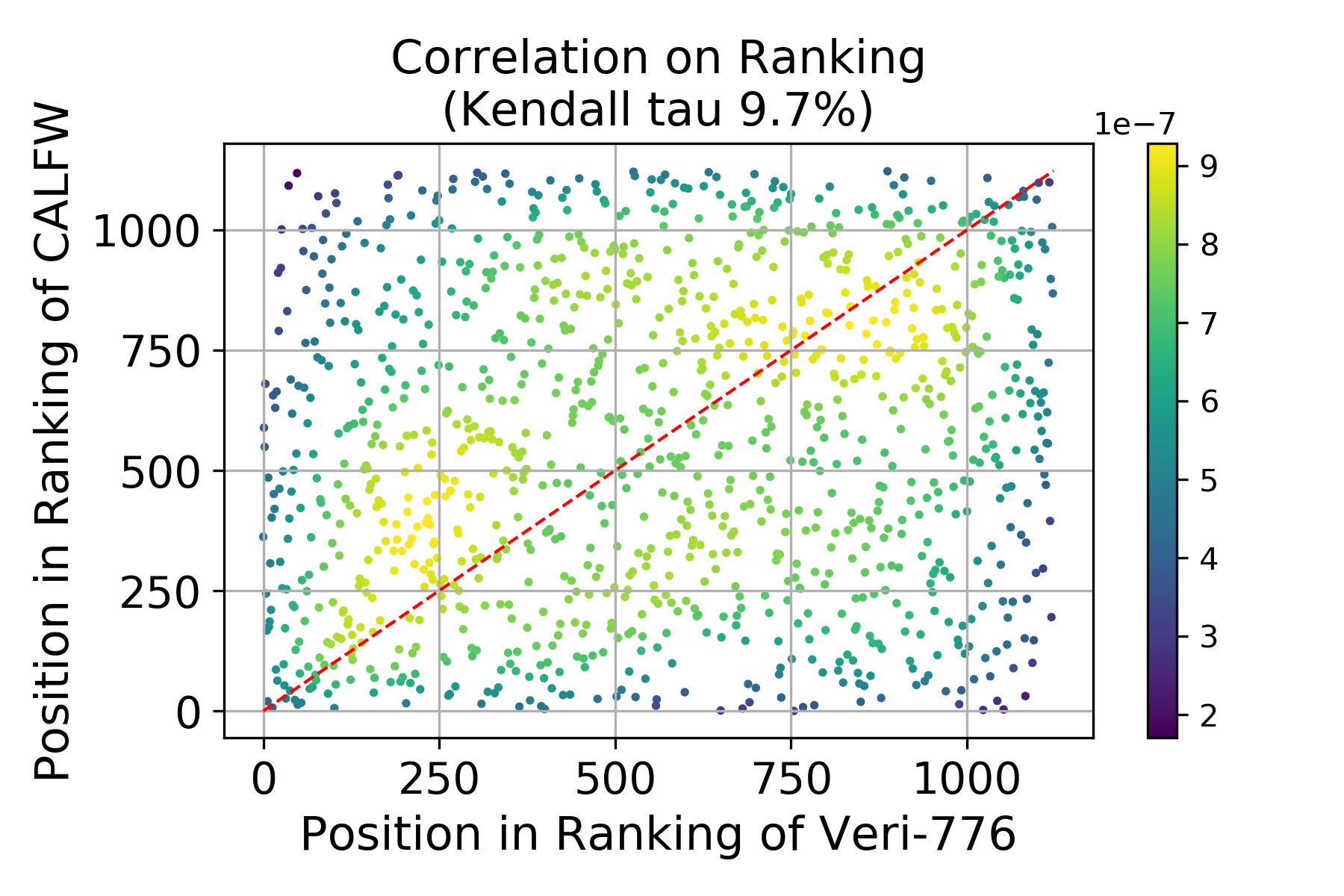}
		\end{minipage}
	}	
	\subfigure[Face and Products]{
		\begin{minipage}[t]{0.48\linewidth}
			\centering
			\includegraphics[width=0.99\linewidth]{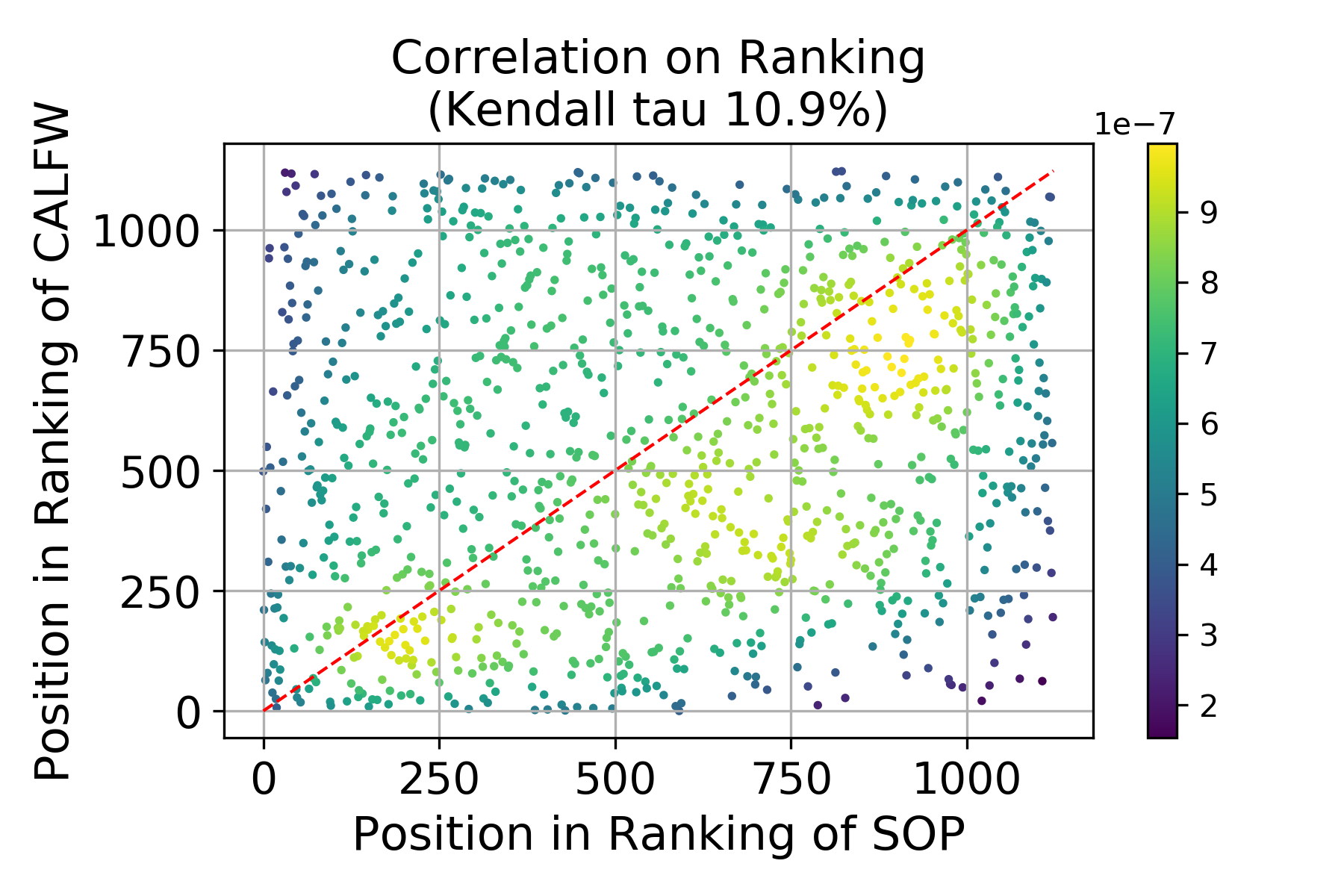}
		\end{minipage}
	}
	\subfigure[Person and Products]{
		\begin{minipage}[t]{0.48\linewidth}
			\centering
			\includegraphics[width=0.99\linewidth]{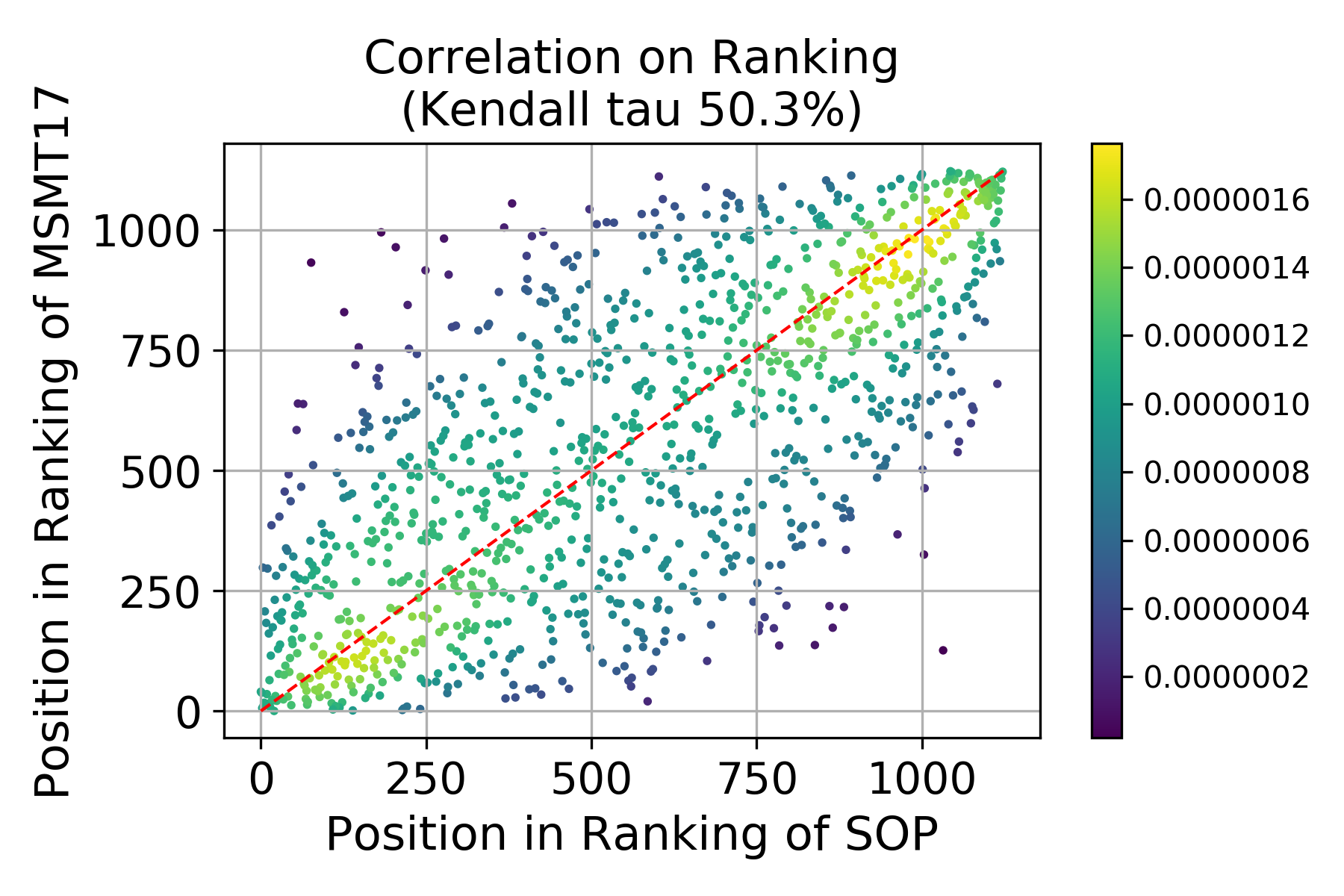}
		\end{minipage}
	}
	\subfigure[Vechicle and Products]{
		\begin{minipage}[t]{0.48\linewidth}
			\centering
			\includegraphics[width=0.99\linewidth]{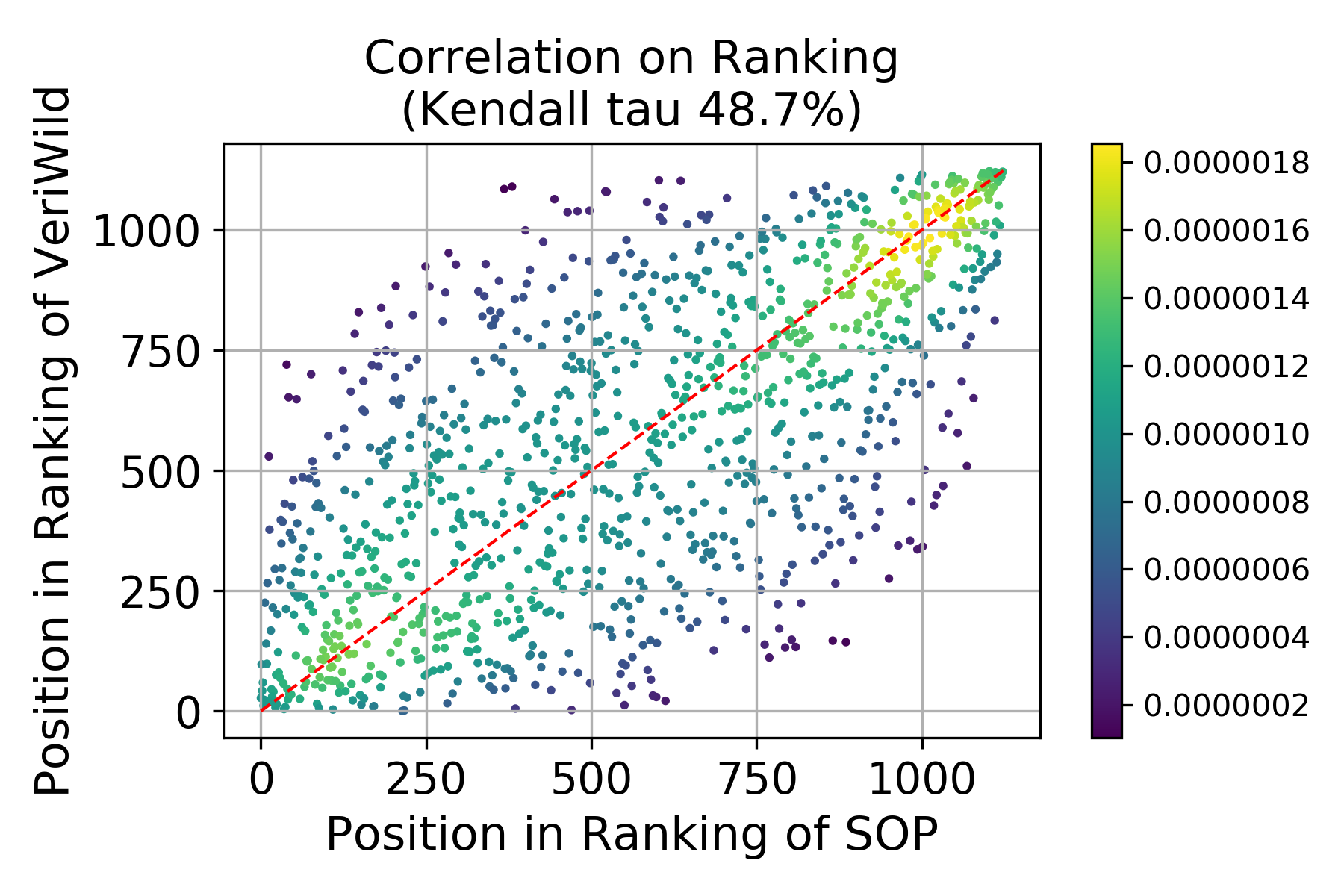}
		\end{minipage}
	} \ 
	\subfigure[Person and Vehicle]{
		\begin{minipage}[t]{0.48\linewidth}
			\centering
			\includegraphics[width=0.99\linewidth]{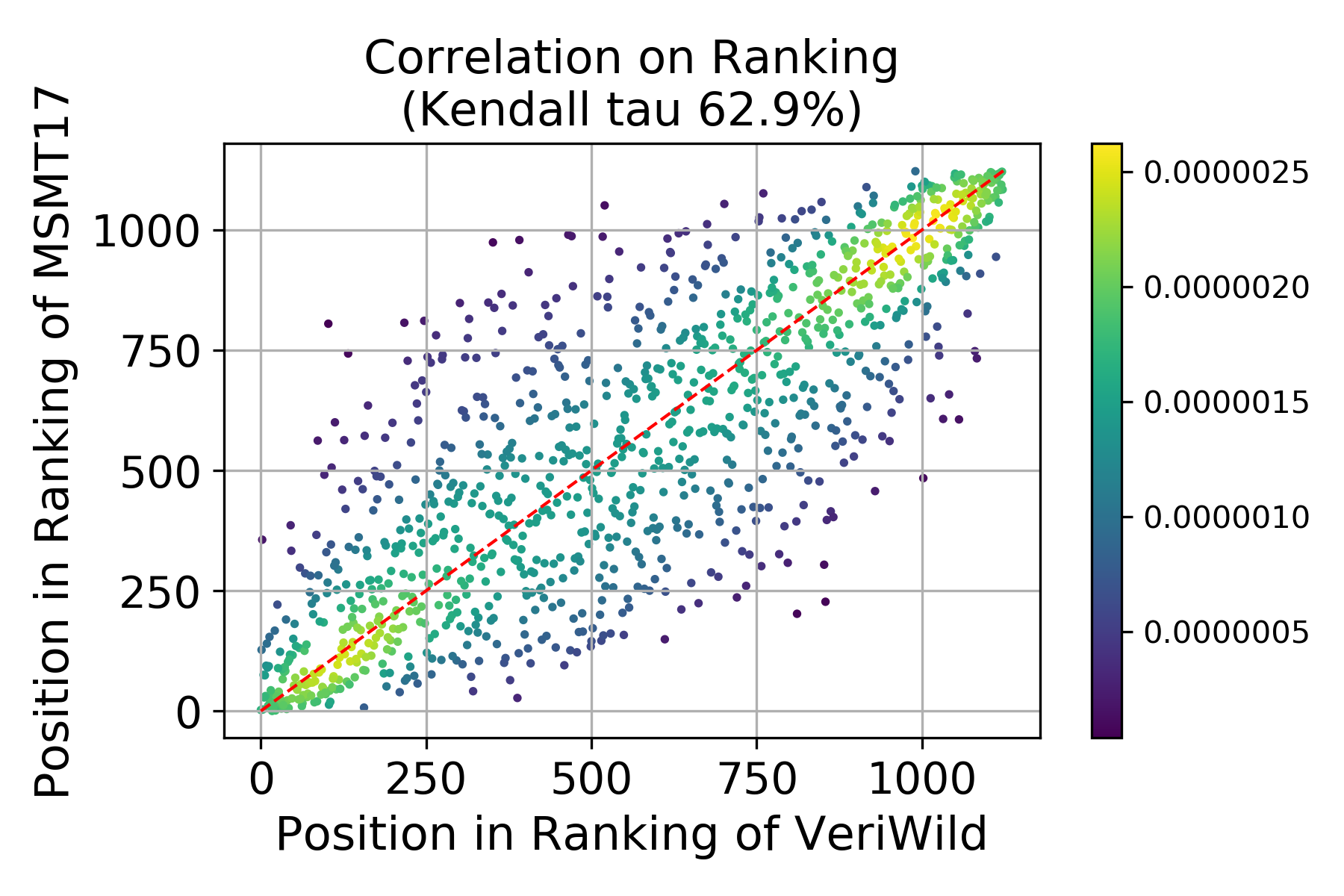}
		\end{minipage}
	}	
	\caption{Selected ranking correlation cross tasks} 
	\label{cross}
	
	\vspace{-5mm}
\end{figure*}

\textbf{Training dataset} MS1M-V3~(MS1M-RetinaFace)~\cite{guo2016ms,deng2019lightweight,deng2020retinaface}, Market1501-Train~\cite{zheng2015person}, MSMT17-Train\cite{wei2018person}, Veri-776-Train~\cite{liu2016deep}, VehicleID-Train~\cite{liu2016deep}, VeriWild-Train~\cite{lou2019veri} and SOP-Train~\cite{oh2016deep} are used as training dataset. The detailed information is shown in table \ref{table:Tranning}. \\
\textbf{Test dataset} Accordingly, we use LFW~\cite{huang2008labeled}, CPLFW~\cite{zheng2018cross}, CFP~\cite{sengupta2016frontal}, CALFW~\cite{zheng2017cross}, AGEDB-30~\cite{moschoglou2017agedb}, Market1501-Test~\cite{zheng2015person}, MSMT17-Test~\cite{wei2018person}, Veri-776-Test~\cite{liu2016deep}, \\ VehicleID-Test~\cite{liu2016deep},  VeriWild-Test~\cite{lou2019veri} and SOP-Test~\cite{oh2016deep} as test dataset. The detailed information is shown in table \ref{table:Test}.\\
\textbf{Seach space} The search space is set as follows: $\mathcal H = \{10, 11, 12\}$, $\mathcal M = \{3, 3.5, 4\}$, $\mathcal T = \{t_1,t_2,t_3,t_4\}$, $\mathcal G = \{g_{share}, g_1,g_2,g_3, g_4\}$ and the drop choices of the first $10$ layers are all set to $1$ \xt{and $\lambda$ is set to $1/4$. The subset size $|\mathcal S|$ is set to 10,0000 and we sample 500 sub networks from $\mathcal S$ for training where each sampled sub network is evaluated on all benchmarks. }  \xt{The sampled networks and $\mathcal S$ are released as one of the first multi-task NAS benchmark and has supported the performance prediction track of the second lightweight NAS challenge of CVPR 2022 ({\url{https://cvpr-nas.com/competition}}).}   \\
\textbf{Sampling strategy} We sample data from four tasks to form a batch, input the batch to the shared transformer backbone network, and finally separate four head networks, each of which is responsible for the output of a task. The four tasks separately calculate the loss and sum it up as the total loss. \\ 
\textbf{Training configurations} Because the input size and model structure used by different tasks are quite different. From the model optimization level, the batch size, learning rate and even the optimizer are all different. In order to facilitate subsequent multi-task training, we first unify the model structure and optimization method of each task. In particular, we use transformer as the backbone network. The unified configurations are shown in table \ref{table:setting_config}.

\xt{Experiments with bigger backbones and with more dataset can be found in} \url{https://github.com/PaddlePaddle/VIMER/tree/main/UFO}.

\begin{figure*}[t!]
\vspace{-6mm}
	\subfigure[Ranking correlation between ground truth and prediction on MSMT17. ]{
		\begin{minipage}[t]{0.48\linewidth}
			\centering
			\includegraphics[width=0.99\linewidth]{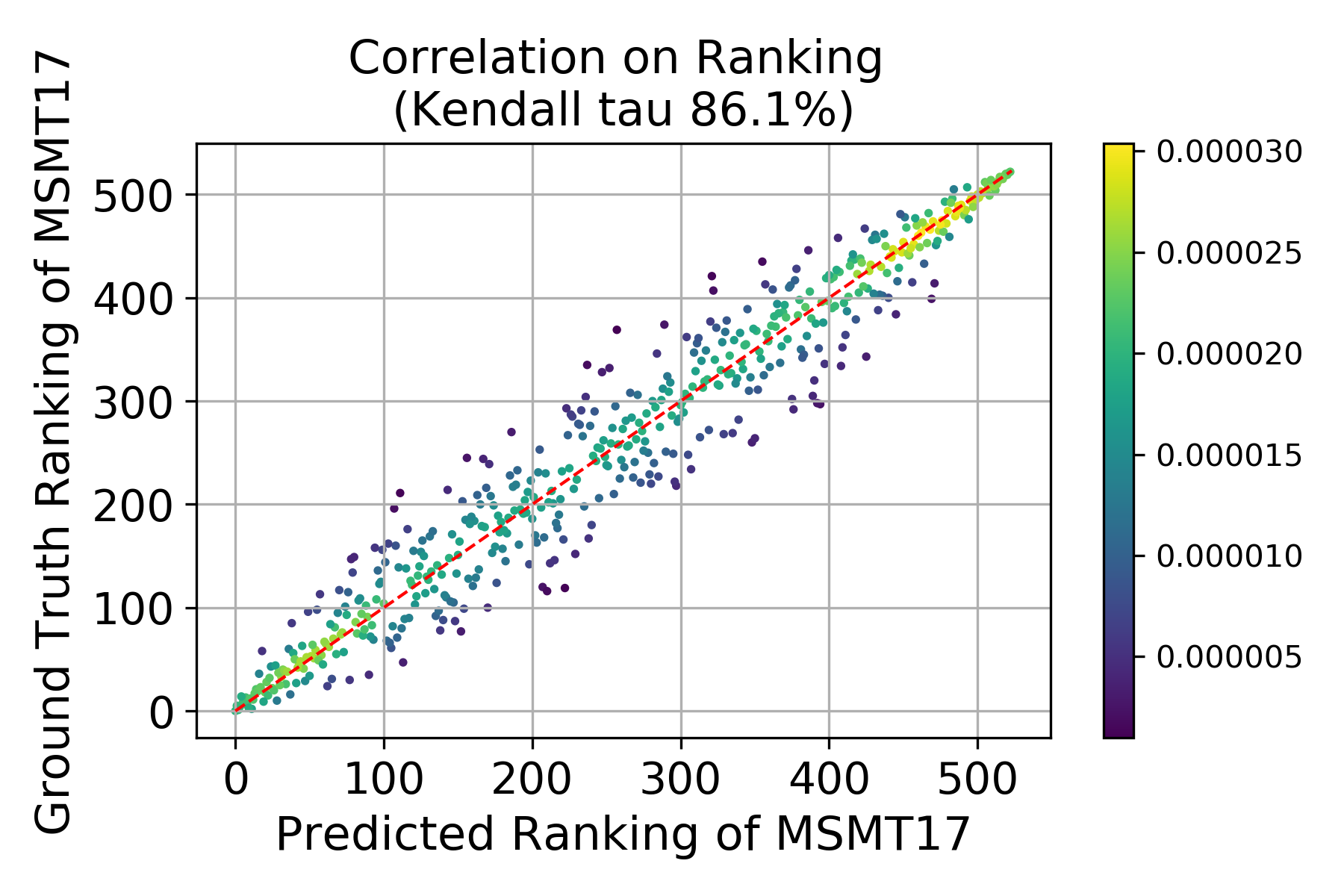}
		\end{minipage}
	}	
	\subfigure[Ranking correlation between ground truth and prediction on SOP.]{
		\begin{minipage}[t]{0.48\linewidth}
			\centering
			\includegraphics[width=0.99\linewidth]{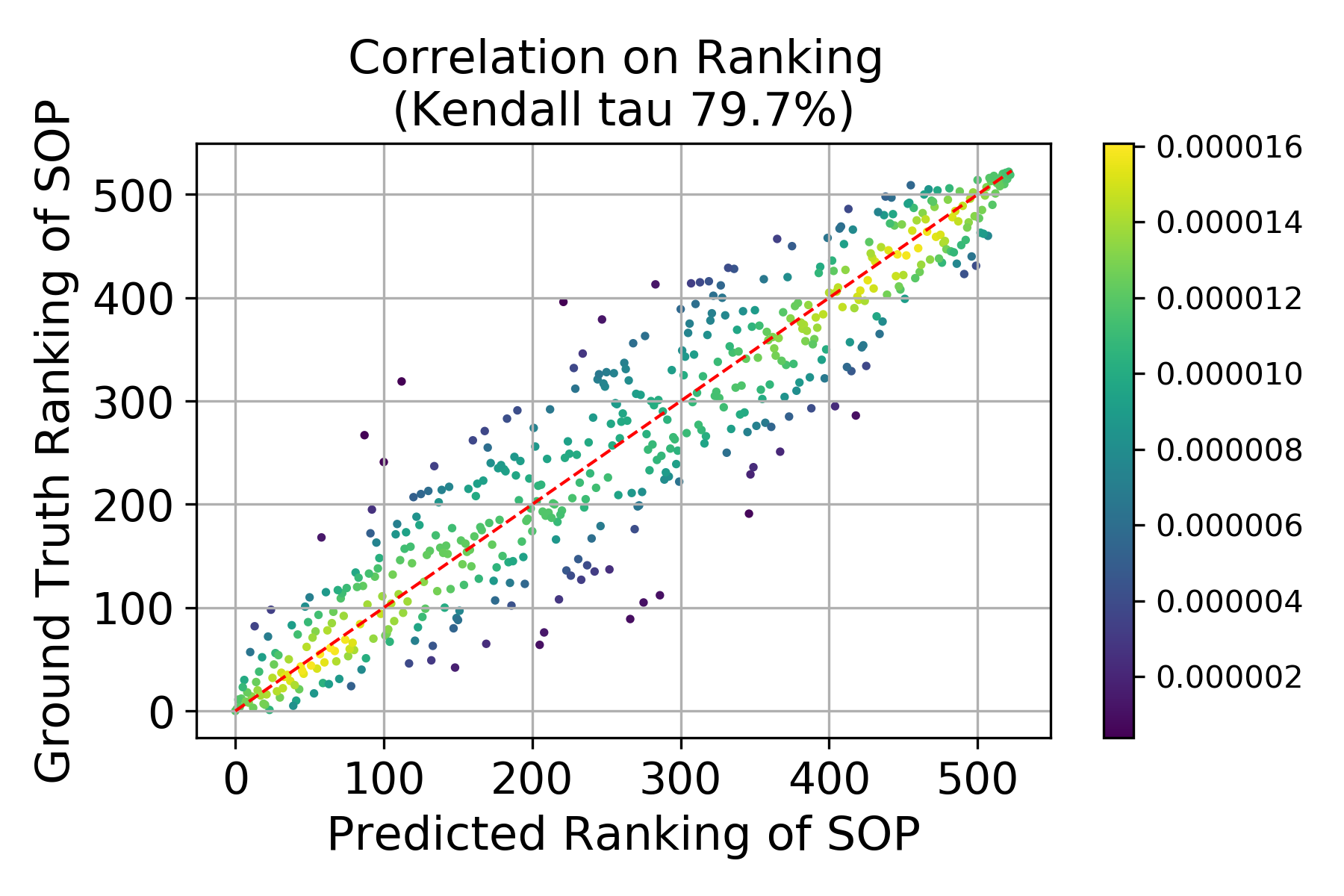}
		\end{minipage}
	}
	\subfigure[Ranking correlation between ground truth and prediction on Market1501.]{
		\begin{minipage}[t]{0.48\linewidth}
			\centering
			\includegraphics[width=0.99\linewidth]{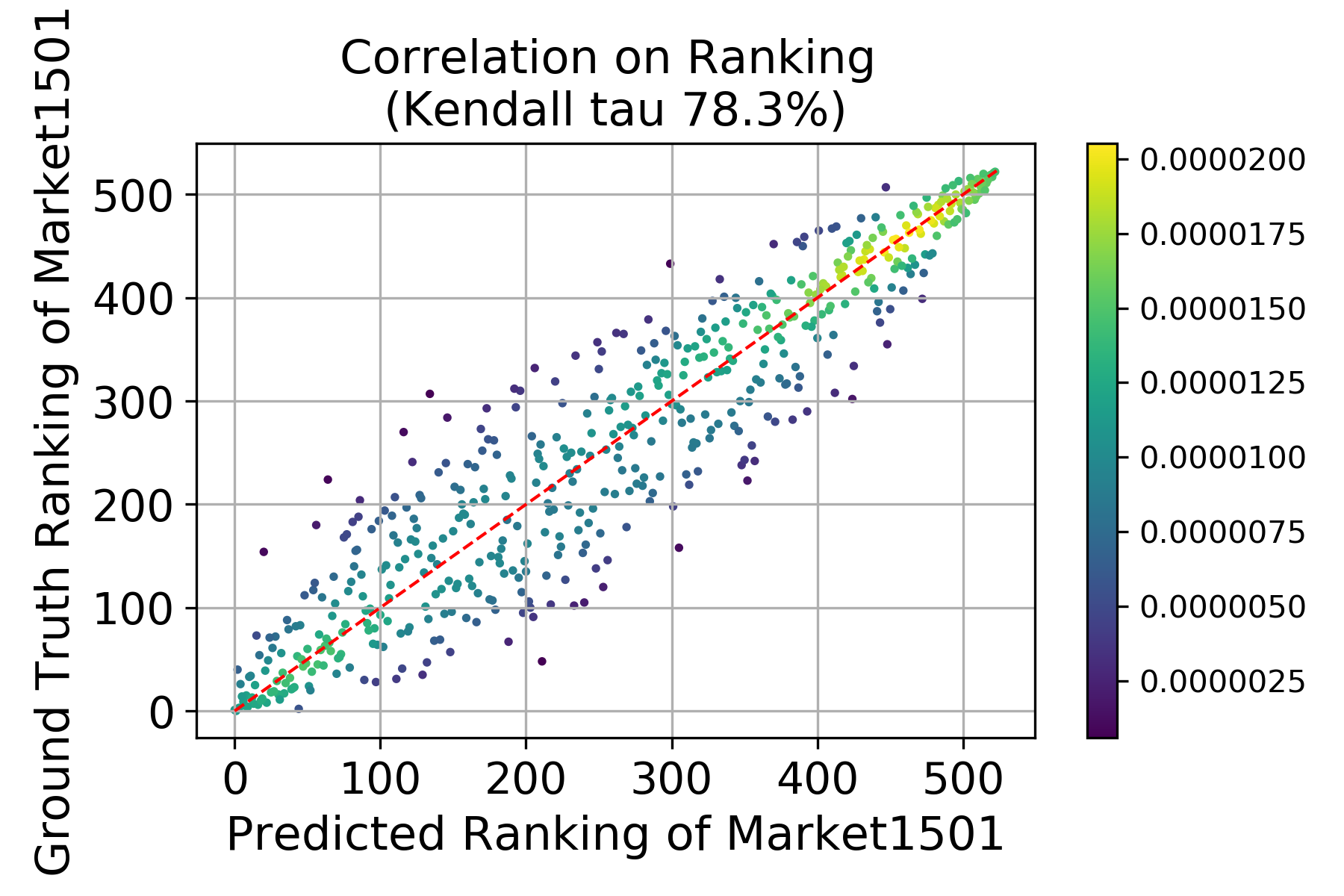}
		\end{minipage}
	} \ 
	\subfigure[Ranking correlation between ground truth and prediction on VeriWild.]{
		\begin{minipage}[t]{0.48\linewidth}
			\centering
			\includegraphics[width=0.99\linewidth]{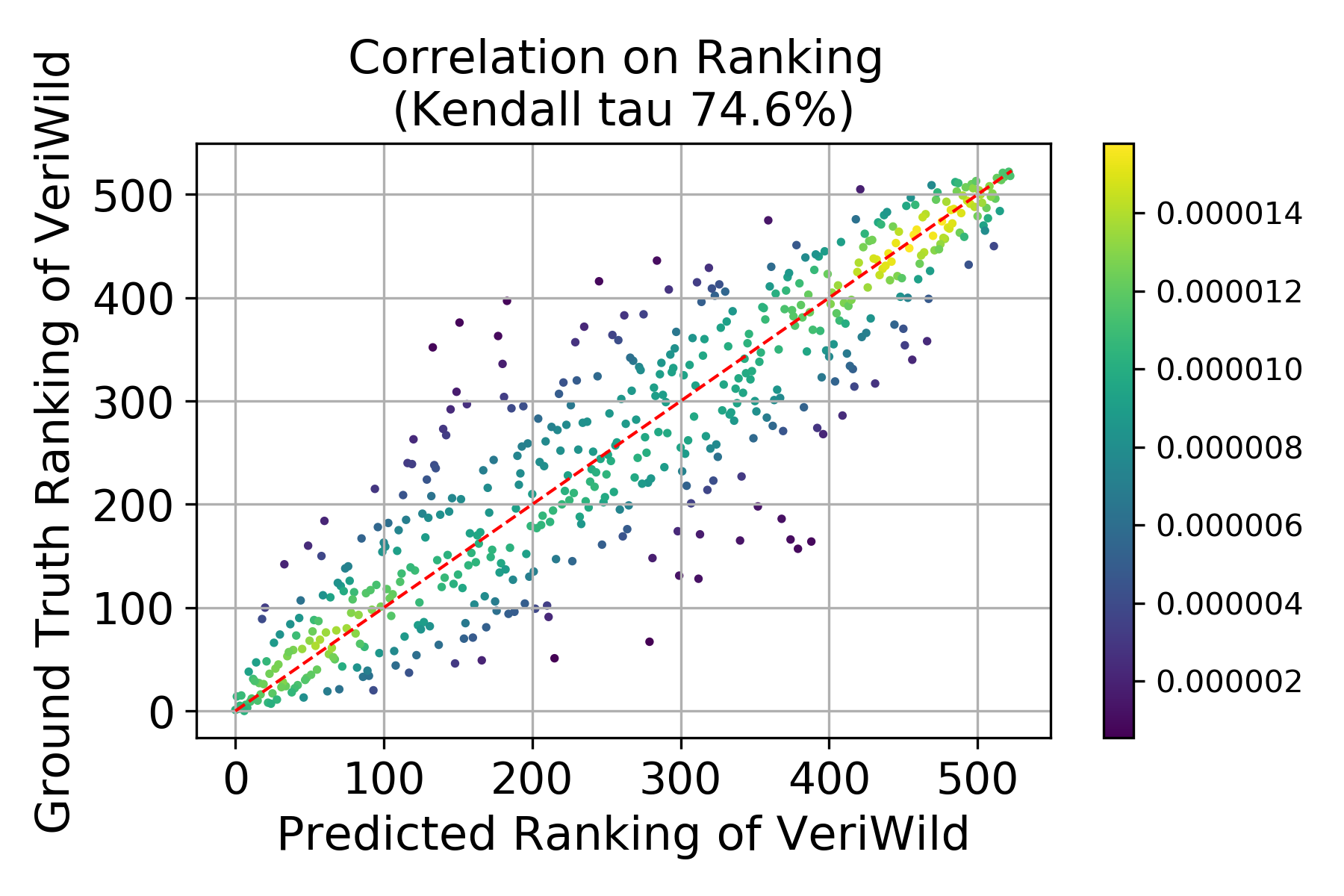}
		\end{minipage}
	}
 	\caption{Performance of task specific predictors on selected benchmarks.}\label{rank_prediction}
	\vspace{-8mm}
\end{figure*}

\subsection{Correlation within and cross tasks}

As different benchmark has different metric, we utilize Kendall tau to measure ranking correlation within and cross tasks. In this experiment, we have sampled 1000+ architectures \xt{from $\mathcal S$} with different reduced parameters and evaluated the performance of each architecture on all benchmarks. 

We first conduct experiment on person benchmarks and vehicle benchmarks. As shown in figure \ref{within}, \xt{person benchmarks are highly correlated while, vehicle benchmark are sightly correlated.} 

Then, we conduct experiments on benchmarks between different tasks.  As shown in figure \ref{cross}, the benchmark of face is sightly correlated to all other tasks. While, certain benchmarks of person, vehicle and products are highly correlated. 

\begin{table*}[t!]
\caption{The impact of FFN paths. }
\vspace{-4mm}
\label{exp:ab_ffn}
\begin{center}
{
\begin{tabular}{cccccccc}
\hline
\multirow{2}{*}{Datasets} & \multirow{2}{*}{All shared} & \multicolumn{2}{c}{UFO-Face} & \multicolumn{2}{c}{UFO-Person}& \multicolumn{2}{c}{UFO-SOP}   \\
\cline{3-4} 
\cline{5-6}
\cline{7-8}
& &  with  & w/o  & with  & w/o & with  & w/o \\
\hline
CALFW   & 95.86    & \textbf{96.00} & 95.95 & \textbf{96.02} & 95.92 & 95.90 & 95.92  \\
\hline
CPLFW   & 93.53    & 93.62 & 93.65 & \textbf{93.62} & 93.37 & 93.30 & 93.37 \\
\hline
Market1501 & 88.19  & \textbf{89.33} & 87.57 & \textbf{89.44} & 87.62 & \textbf{89.44} & 87.62   \\
\hline
MSMT17  &   60.08 & \textbf{64.24} & 61.13  & \textbf{64.84} & 61.38 & \textbf{64.66} & 61.38\\
\hline
Veri-776 &  86.11    & \textbf{87.90} & 86.48   & \textbf{88.03} & 86.50& \textbf{87.95} & 86.50  \\
\hline
VehicleID  &86.27  &\textbf{85.43} & 85.40  & \textbf{87.00} & 85.46& \textbf{86.41} & 85.46  \\
\hline
VeriWild  & 68.43&  \textbf{69.60} & 67.85 & \textbf{69.72} & 67.92 & \textbf{69.69}&  67.92 \\
\hline
SOP  & 86.64   & \textbf{86.09} & 83.47 & \textbf{86.19} & 83.50  & \textbf{86.24} & 83.53 \\
\hline
\makecell[c]{Flops reduction\\relative to ViT-base}  & 0\%   & 29.52\% & 27.38\% & 26.94\% & 27.38\% & 27.15\% &  27.38\% \\
\hline
\makecell[c]{Param reduction\\relative to ViT-base}  & 0\%   & 29.39\% & 27.25\% & 26.88\% & 27.31\% & 26.88\% &  27.10\%\\
\hline
\end{tabular}
}
\end{center}
\vspace{-6mm}
\end{table*}

\subsection{The performances of task specific predictors} 
  
\xt{Figure \ref{rank_prediction} illustrates the performance of task specific predictors. \xt{In the experiment, we sample 500 sub networks for training and 500+ sub networks (no overlap) for testing where each sampled sub network is evaluated on all tasks. }  Then, we measure the correlation between the predicted rankings and ground truth rankings. As shown in the figure, the predictors are with very good accuracy.}

\subsection{The impact of FFN paths} 

\xt{As shown in \textcolor{black}{table \ref{exp:ab_ffn}}, FFN paths is of significant importance in multi-task learning, compared with all shared approach. The supernet with FFN paths can significantly improve the performances. One possible explanation is that, FFN paths relieve the competition of shared attention weights. }\\

\begin{table*}[t!]
\caption{Compared with SOTA results}
\vspace{-6mm}
\label{exp:sota}
\begin{center}
{
\begin{tabular}{cccccccc}
\hline
 & SOTA 
 & \makecell[c]{UFO for \\  CPLFW}
 & \makecell[c]{UFO for \\  CFP-FF}
 & \makecell[c]{UFO for \\  Veri776}
 & \makecell[c]{UFO for \\  VehicleID}
 & \makecell[c]{UFO for \\  SOP} \\
\hline
CALFW   & 96.20    & 95.78 & 95.98  &95.95 &95.83 & 95.95  \\
\hline
CPLFW   & 93.37    & 93.65 & 93.40 &93.60 &93.47 & 93.63  \\
\hline
CFP-FF  & 99.89   & 99.87 & 99.89   &99.83 &99.86 & 99.86  \\
\hline
Market1501 & 91.50  & 89.45 & 89.52  &89.46 &89.51 & 89.50  \\
\hline
MSMT17  &   69.40 & 64.84 & 64.86   &64.87 &65.03 & 65.06 \\
\hline
Veri-776 &  87.10 & 88.03 & 88.01   &88.18 &88.09 & 88.06  \\
\hline
VehicleID &80.50  &85.26 & 86.32  &86.18 &87.04 & 86.44  \\
\hline
VeriWild  & 77.30&  69.74 & 69.71  &69.73 &69.82 & 69.74  \\
\hline
SOP  & 85.90  & 86.19 & 86.22  &86.21 &86.25 & 86.39  \\

\hline
\makecell[c]{Flops reduction\\relative to ViT-base} & -   & 18.90\% & 25.00\%  & 11.50\% & 20.00\% &  24.00\% \\
\hline
\makecell[c]{Param reduction\\relative to ViT-base}  & -   & 17.60\% & 23.90\% & 10.30\% & 18.90\% &  22.80\%\\
\hline
\makecell[c]{Flops reduction\\relative to supernet}  & -   & 50.84\% & 54.54\% & 46.36\% & 51.51\% & 53.93\% \\
\hline
\makecell[c]{Param reduction\\relative to supernet}  & -   & 41.97\% & 46.40\%  & 36.83\%  & 42.88\% & 45.63\% \\
\hline
\end{tabular}
}
\end{center}
\vspace{-6mm}
\end{table*}    
    
\subsection{Compared with SOTA results}

In this subsection, we compare UFO with previous SOTA results \footnote{ without rerank strategy and external data from https://paperswithcode.com/} on 10 benchmarks, that are SOTA results on CALFW~\cite{an2021partial}, CPLFW~\cite{kim2020groupface}, CFP-FF~ \cite{chrysos2021deep}, Market1501~\cite{herzog2021lightweight}, MSMT17~\cite{he2021transreid}, Veri-776~\cite{huynh2021strong}, VehicleID~\cite{he2020fastreid}, VeriWild~\cite{he2020fastreid} and SOP~\cite{lee2020compounding}. As shown in \textcolor{black}{table \ref{exp:sota}}, UFO reaches SOTA result on CFP-FF and creates 4 new SOTA results on CPLFW, Veri-776, VehicleID and SOP. We can averagely reduce \textcolor{black}{51.08\% flops and 42.34\%} parameters (relative to supernet). \xt{ Besides,  we also compare UFO against two recent state-of-the-art multi-task methods, \emph{i.e.}, Switch Transformers \cite{fedus2021switch}  and DSelect-k \cite{hazimeh2021dselect} (based on their publicly available code).  Our UFO surpasses Switch Transformers/DSelect-k by 1.28/1.05\% (CALFW), 1.61/1.30\% (CPLFW), 0.29/0.25\% (CFP-FF), 1.43/0.98\%  (Market-1501), 1.79/2.18\%  (MSMT17), 5.39/6.27\% (Veri-776),  6.28/6.47\% (VehicleID),  12.32/12.57\%  (VeriWild),  0.03/0.62\% (SOP), respectively.}

\section{Conclusions}
This paper proposes a novel train-and-deploy paradigm named Unified Feature Optimization (UFO) to benefit down-stream tasks with large-scale pretraining. UFO maintains the benefit of large-scale pretraining and provides high convenience for flexible deployment: it transfers the multi-task trained supernet to a dedicated model for any already-seen sub-tasks without no adaptation cost and reduces the model size during adaptation. On deep representation learning tasks, we explore an early prototype of UFO based on Vision Transformer (ViT) and Neural Architecture Search (NAS) techniques. Specifically, UFO integrates multiple trimming strategies to enhance the trimming flexibility for ViT and employs a novel \textcolor{black}{performance-ranking based method for NAS}. Experimental results show that the sub-model trimmed from supernet surpasses its single-task-trained counterpart and has smaller model size. That being said, we also note that the accuracy improvement and the size reduction is still marginal and call for more efforts from the research community to explore UFO. \xt{Besides, UFO also supported the release of \href{https://github.com/PaddlePaddle/VIMER/tree/main/UFO}{17 billion parameters computer vision (CV) foundation model} which is the largest CV model in the industry.}  

\appendix
\section{Multi-task NAS benchmark}

The sampled sub networks in UFO are released as one of the first multi-task NAS benchmark and has supported the performance prediction track of the second lightweight NAS challenge of CVPR 2022 ({\url{https://cvpr-nas.com/competition}}). 
The Multi-task NAS benchmark is released in \href{https://aistudio.baidu.com/aistudio/datasetdetail/134077?lang=en}{Baidu AI Studio} which is a one-stop developer platform based on Baidu's deep learning platform PaddlePaddle. Baidu AI Studio provides free online courses, free computing power support and non-stop competitions to encourage the development of deep learning.

\section{More Experiments}
 UFO also utilized bigger backbone and more dataset and released the \href{https://github.com/PaddlePaddle/VIMER/tree/main/UFO}{VIMER-UFO}, a task-MoE based 17 billion parameters computer vision foundation model, which supports extraction of lightweight models by sparse activation and achieves SOTA on 28 datasets across a battery of visual recognition tasks.

\begin{figure}[h!]
\vspace{-6mm}
\centering
 \scalebox{0.45}{
\includegraphics{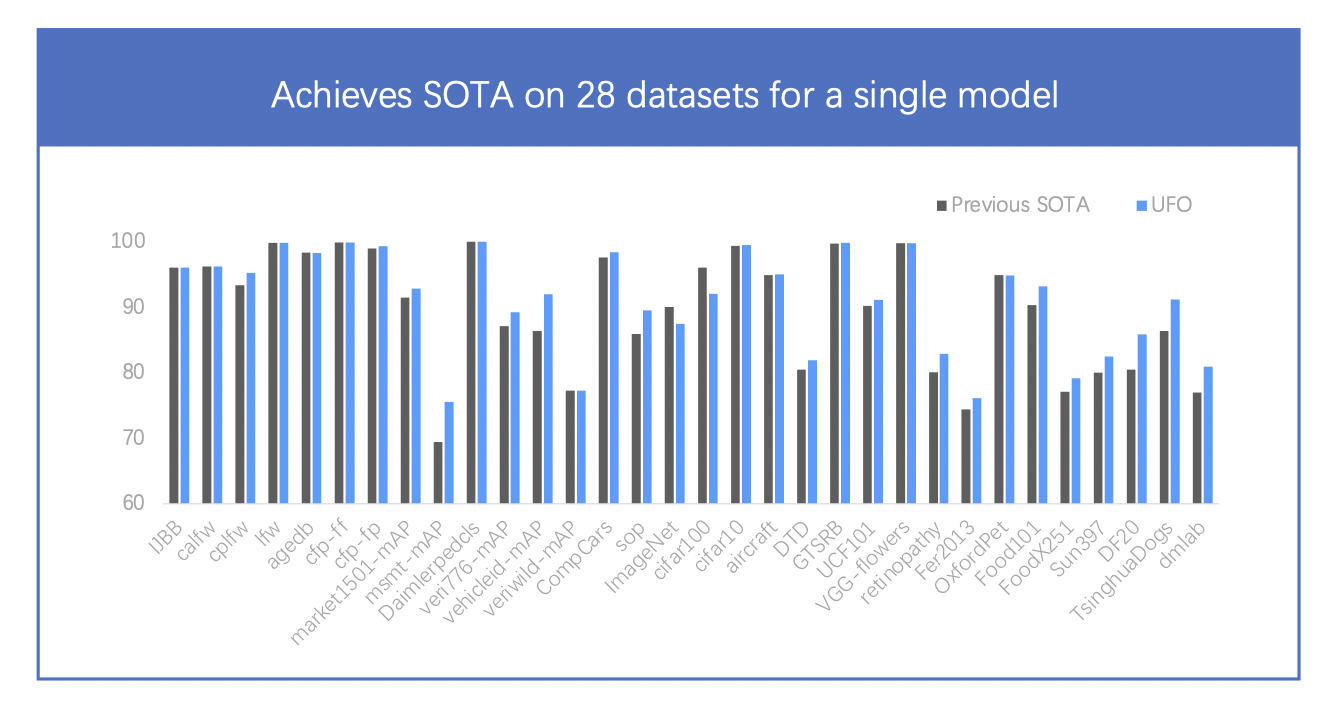}
}
\caption{VIMER-UFO achieves SOTA on 28 datasets for a single model. 
}
\label{fig:sota}
\vspace{-6mm}
\end{figure}

\begin{figure*}[t!] 
	\subfigure[Face and Person ]{
		\begin{minipage}[t]{0.48\linewidth}
			\centering
			\includegraphics[width=0.99\linewidth]{image/Rank_correlation-CALFW-DukeMTMC-10.6.png}
		\end{minipage}
	}
	\subfigure[Face and Vehicle]{
		\begin{minipage}[t]{0.48\linewidth}
			\centering
			\includegraphics[width=0.99\linewidth]{image/Rank_correlation-CALFW-Veri-776-9.7.png}
		\end{minipage}
	}	
	\subfigure[Face and Person]{
		\begin{minipage}[t]{0.48\linewidth}
			\centering
			\includegraphics[width=0.99\linewidth]{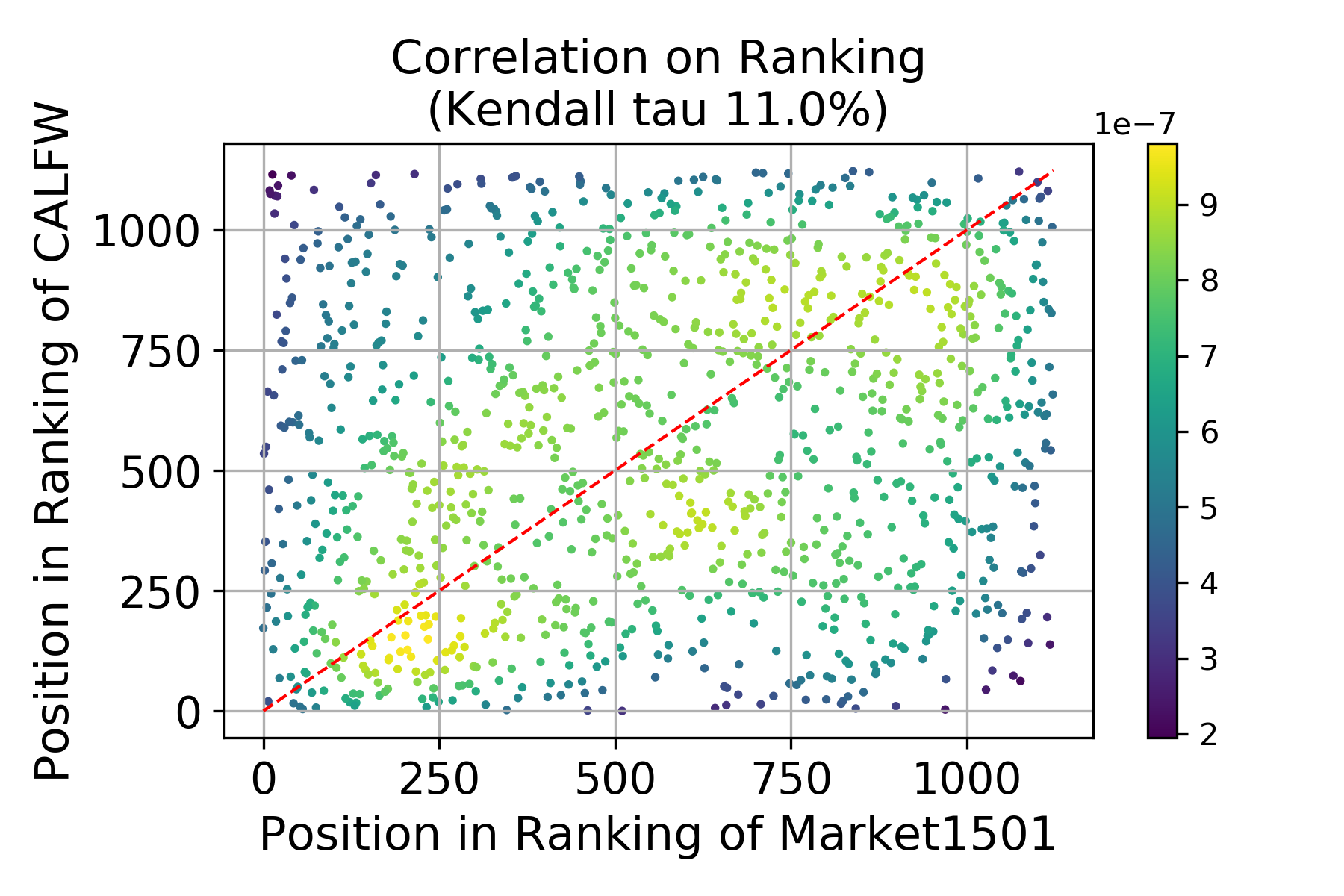}
		\end{minipage}
	}
		\subfigure[Face and Person]{
		\begin{minipage}[t]{0.48\linewidth}
			\centering
			\includegraphics[width=0.99\linewidth]{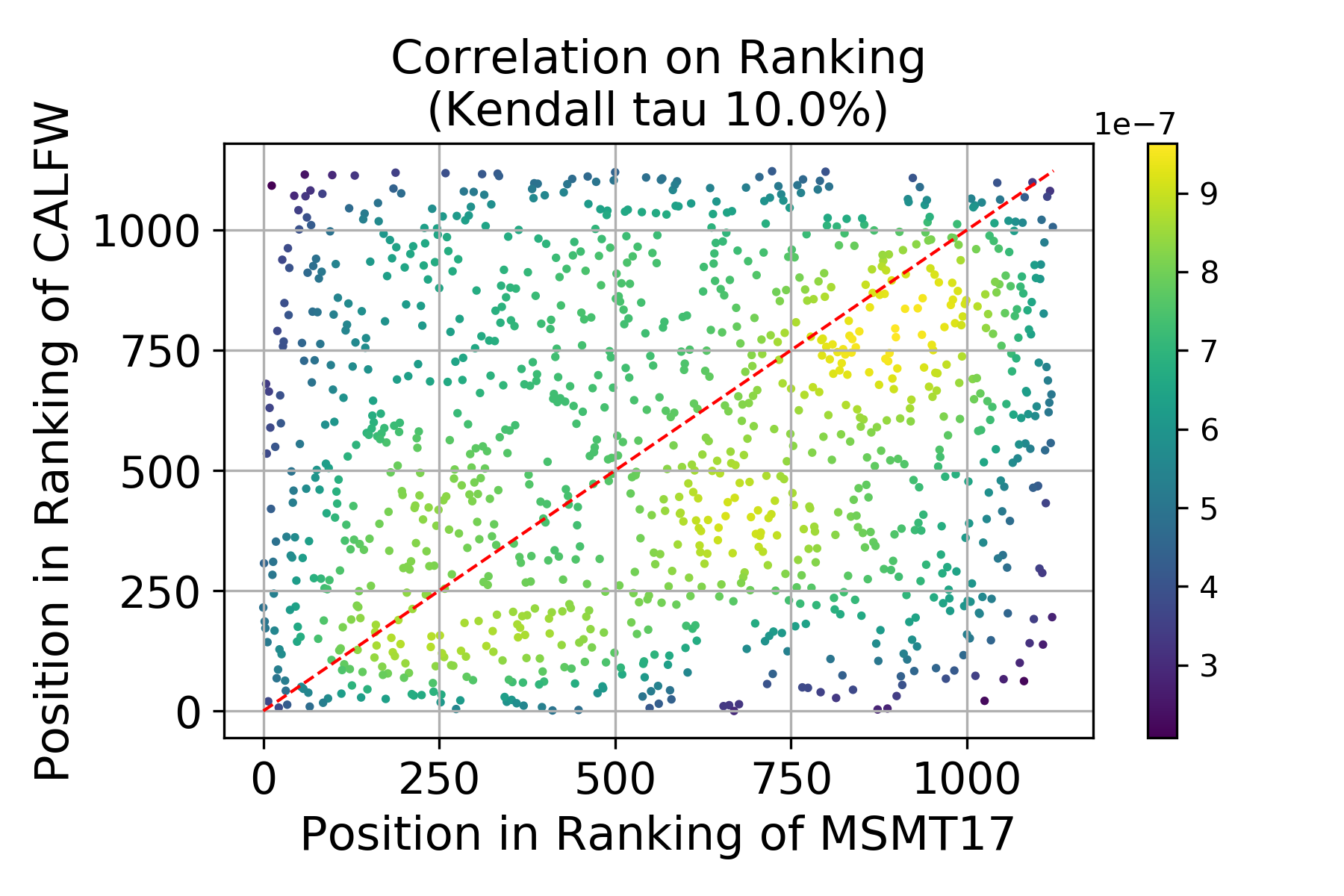}
		\end{minipage}
	}
	\subfigure[Face and Products]{
		\begin{minipage}[t]{0.48\linewidth}
			\centering
			\includegraphics[width=0.99\linewidth]{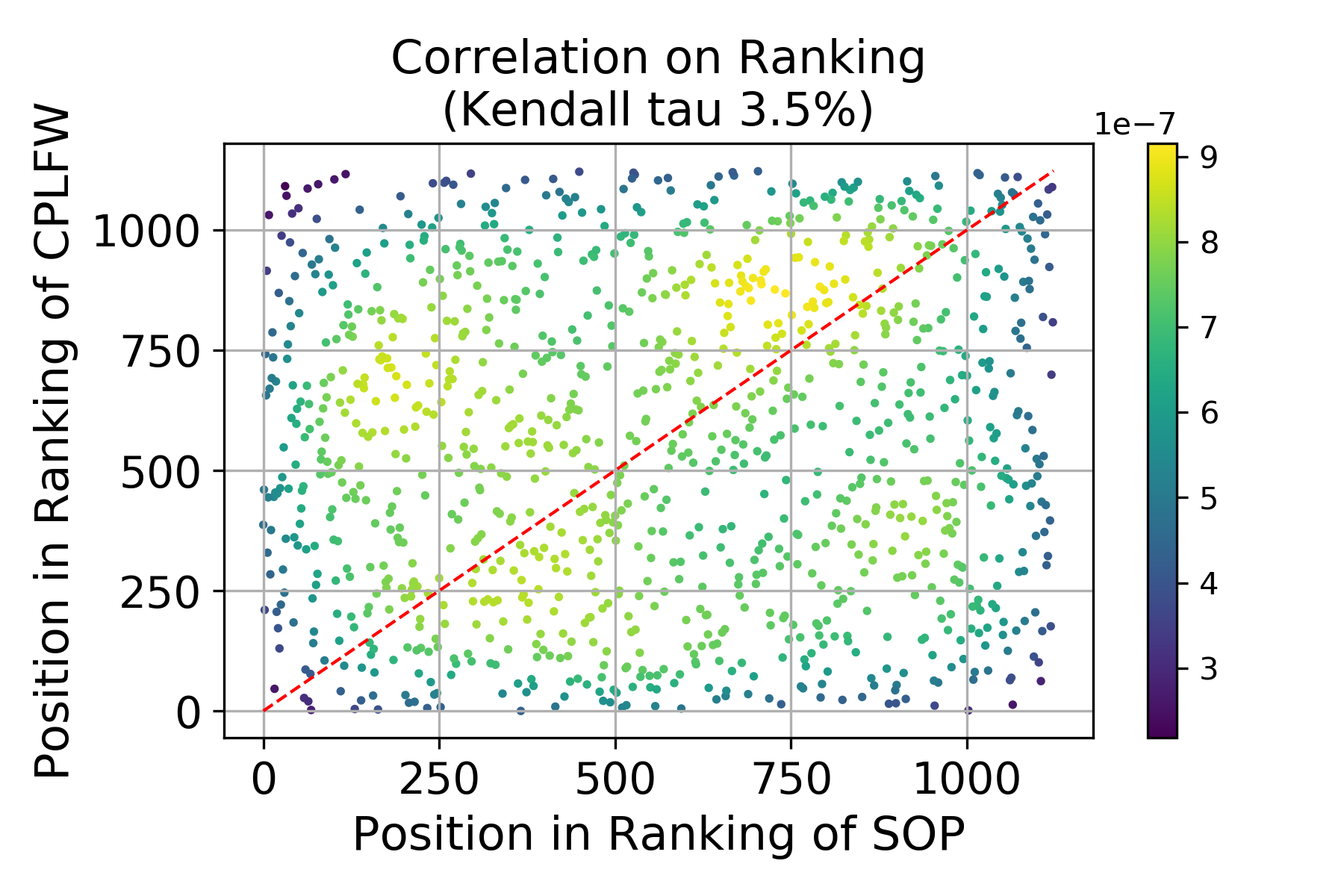}
		\end{minipage}
	}
	\subfigure[Person and Products]{
		\begin{minipage}[t]{0.48\linewidth}
			\centering
			\includegraphics[width=0.99\linewidth]{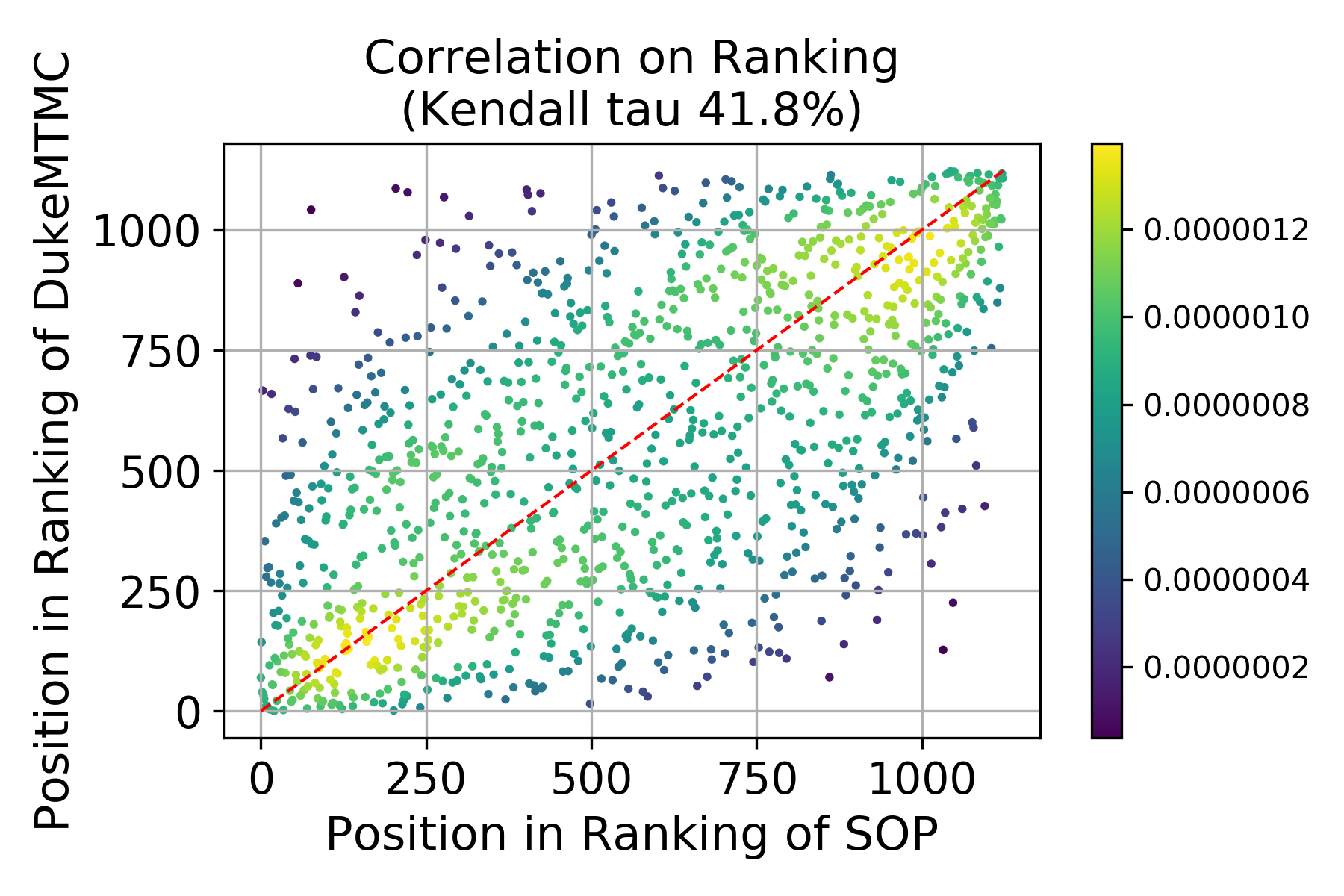}
		\end{minipage}
	}
	\subfigure[Vechicle and Products]{
		\begin{minipage}[t]{0.48\linewidth}
			\centering
			\includegraphics[width=0.99\linewidth]{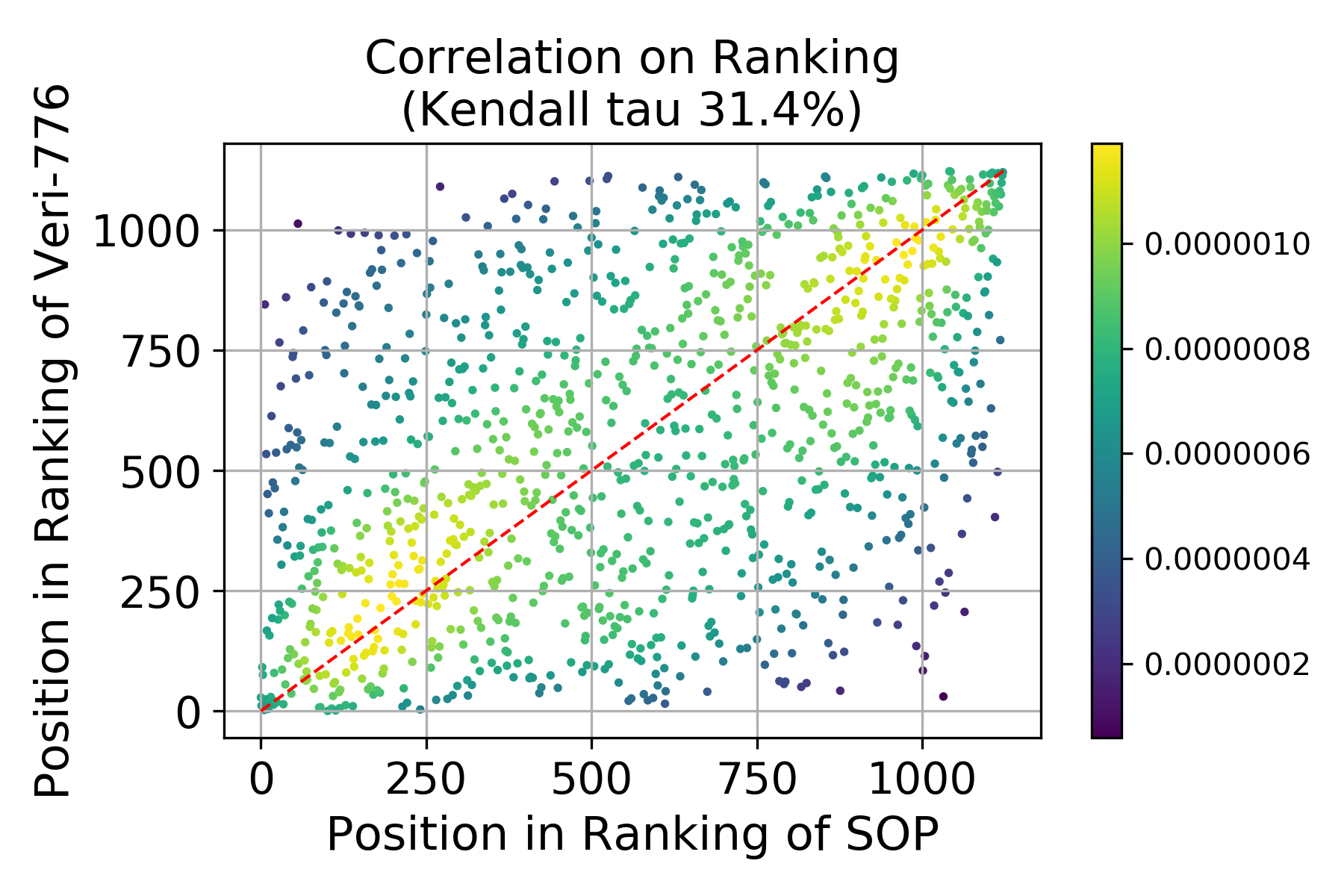}
		\end{minipage}
	} \ 
	\subfigure[Person and Vehicle]{
		\begin{minipage}[t]{0.48\linewidth}
			\centering
			\includegraphics[width=0.99\linewidth]{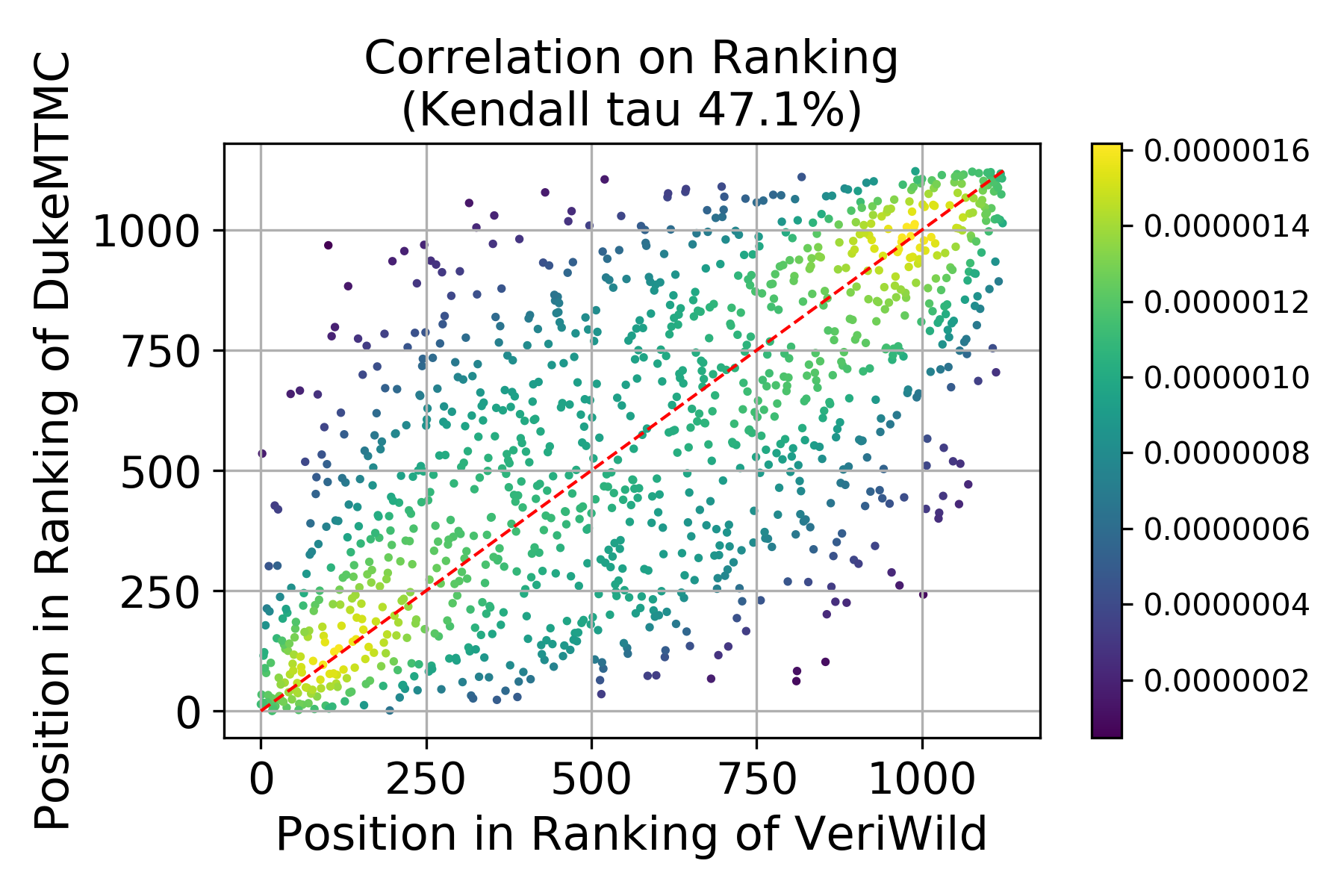}
		\end{minipage}
	}	
	\caption{All cross tasks correlations part one.}
	\label{cross_1}
\end{figure*}

\begin{figure*}[h!] 
	\subfigure[Face and Person ]{
		\begin{minipage}[t]{0.48\linewidth}
			\centering
			\includegraphics[width=0.99\linewidth]{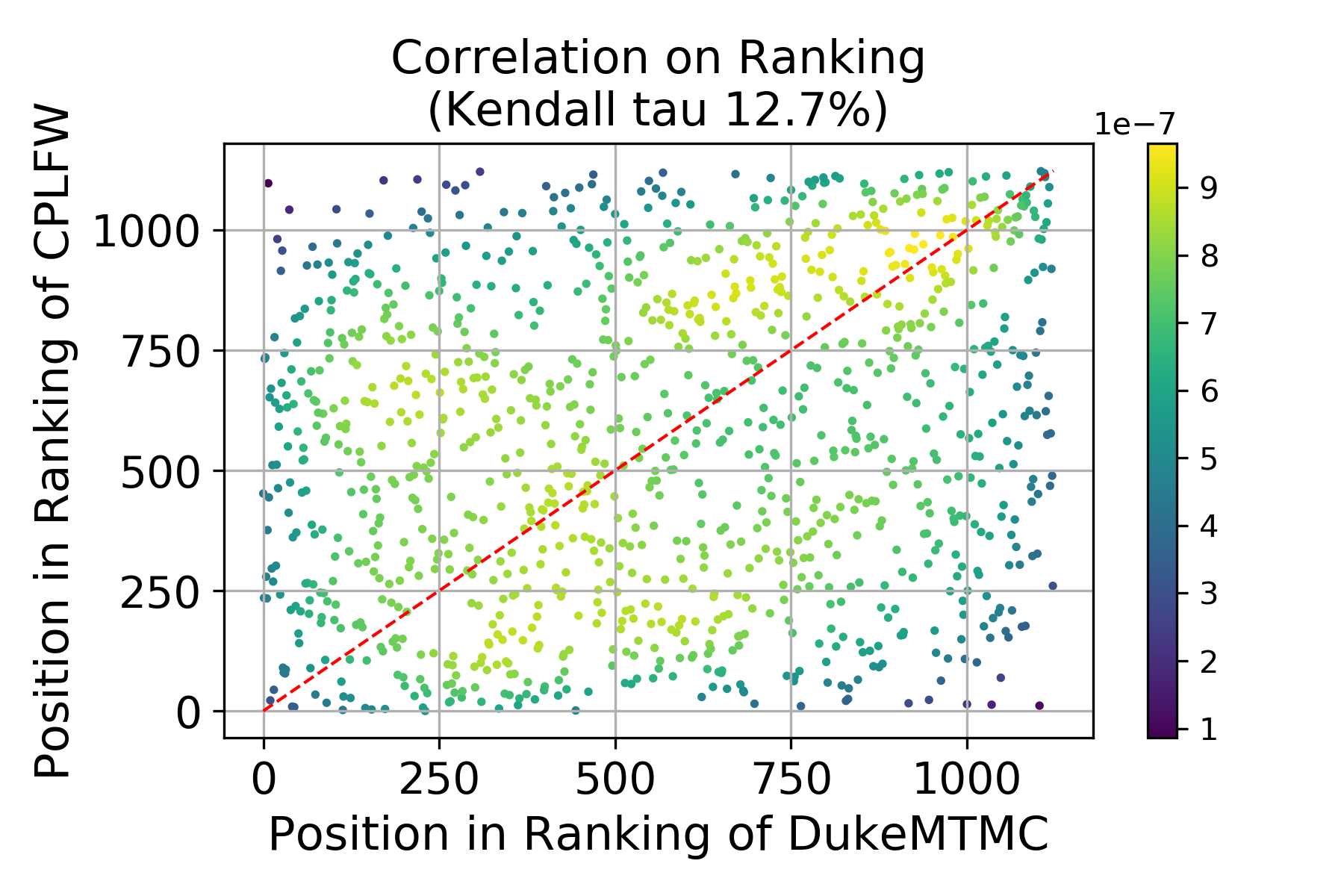}
		\end{minipage}
	}
	\subfigure[Face and Vehicle]{
		\begin{minipage}[t]{0.48\linewidth}
			\centering
			\includegraphics[width=0.99\linewidth]{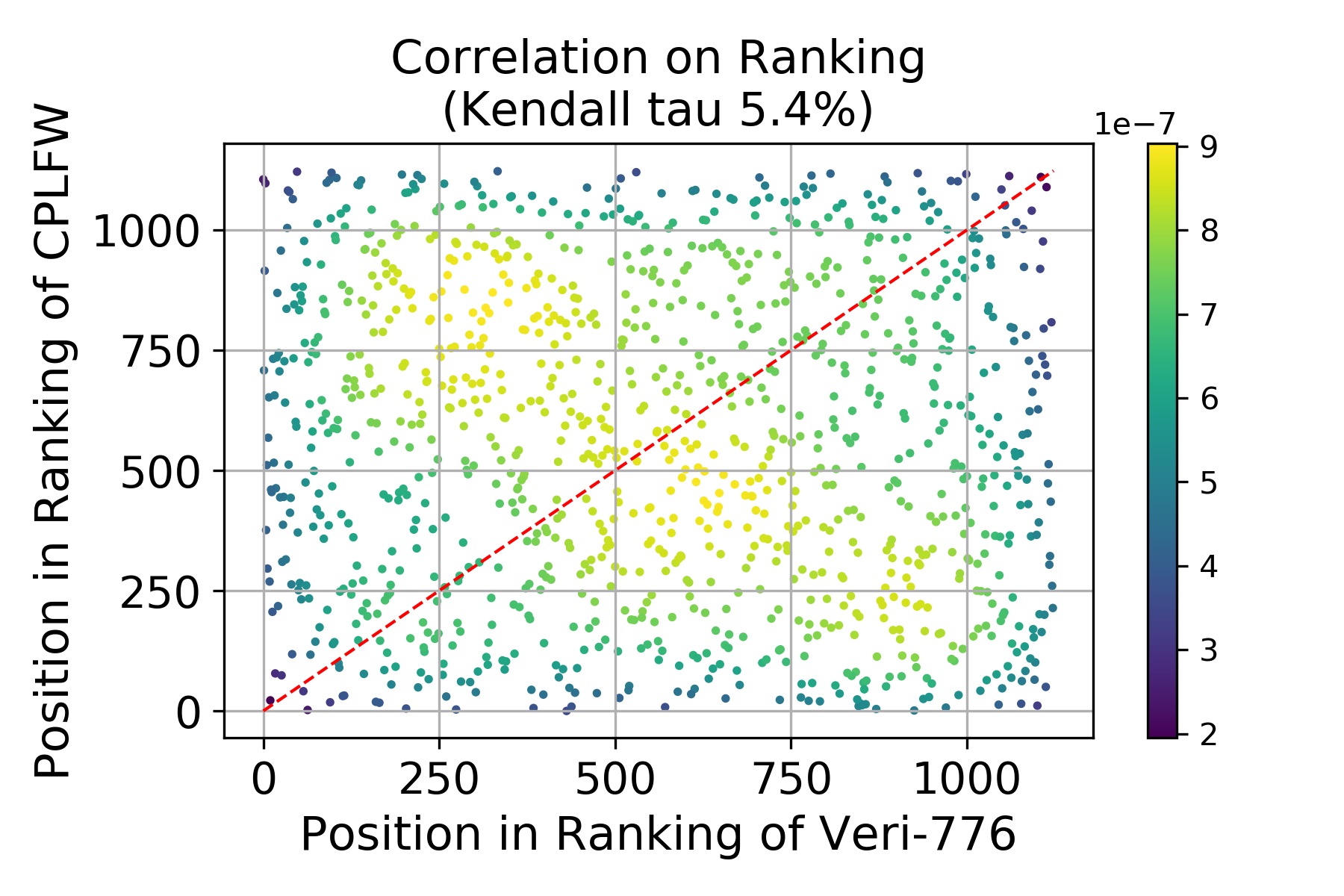}
		\end{minipage}
	}	
	\subfigure[Face and Vehicle]{
		\begin{minipage}[t]{0.48\linewidth}
			\centering
			\includegraphics[width=0.99\linewidth]{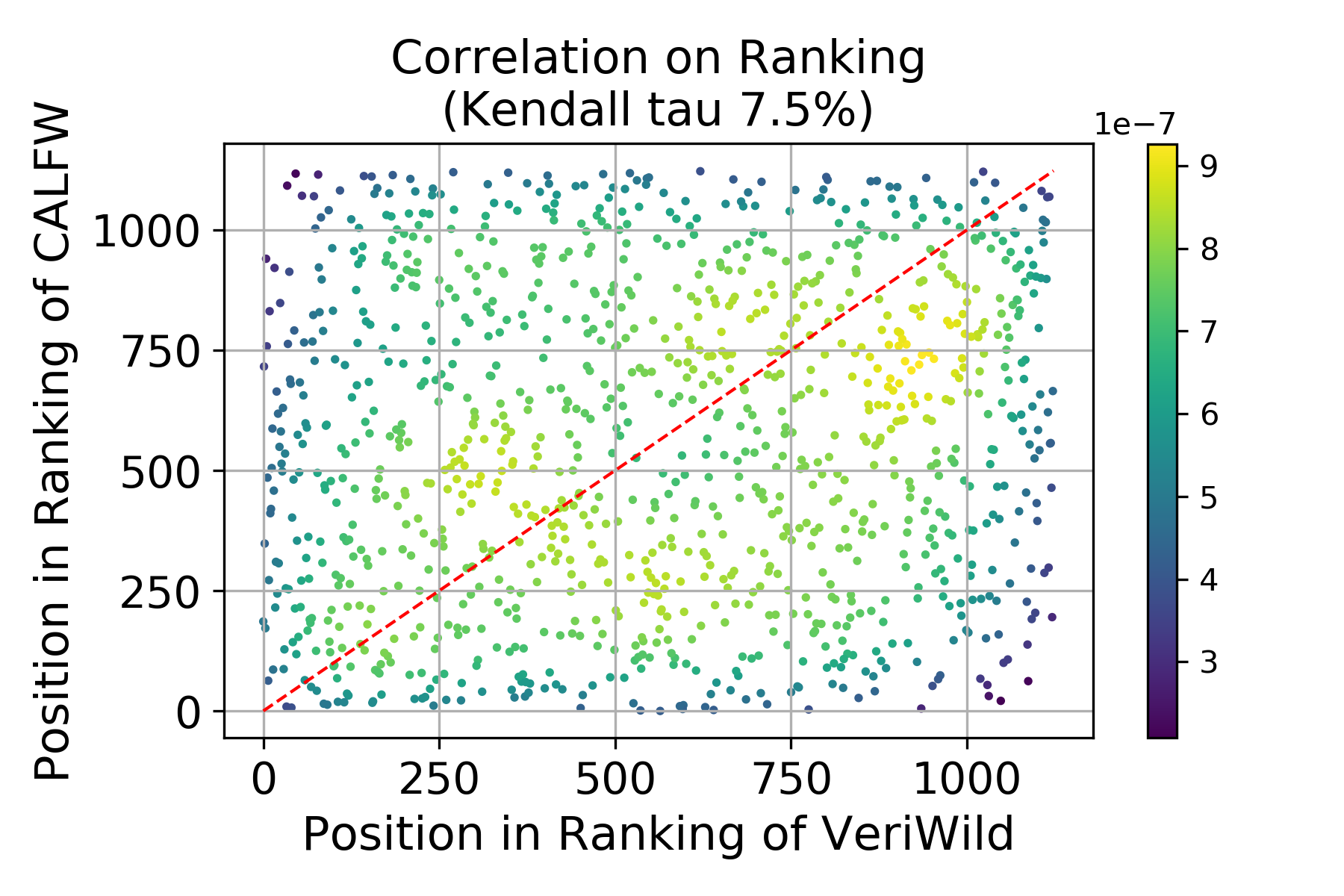}
		\end{minipage}
	}
		\subfigure[Face and Person]{
		\begin{minipage}[t]{0.48\linewidth}
			\centering
			\includegraphics[width=0.99\linewidth]{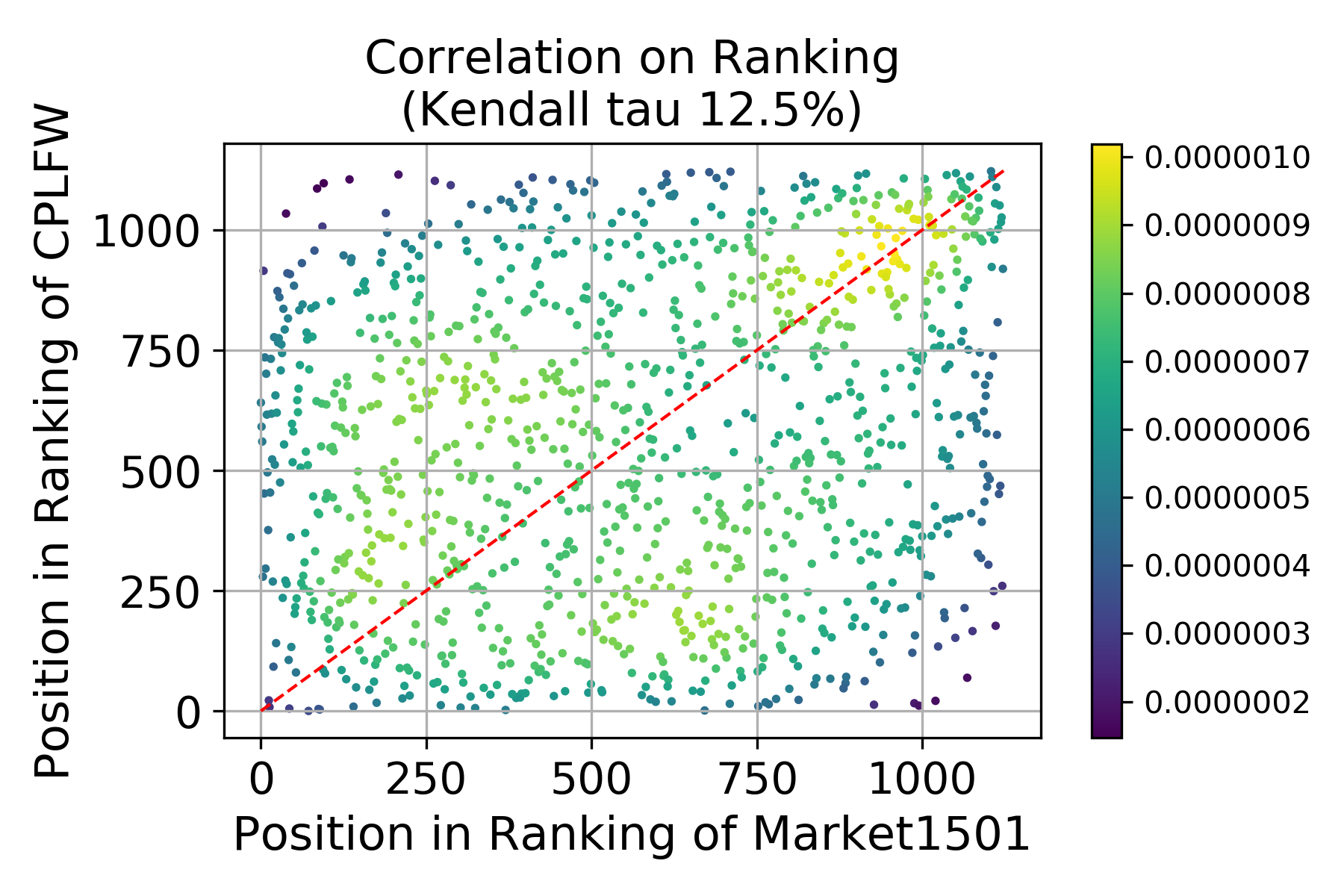}
		\end{minipage}
	}
	\subfigure[Face and Products]{
		\begin{minipage}[t]{0.48\linewidth}
			\centering
			\includegraphics[width=0.99\linewidth]{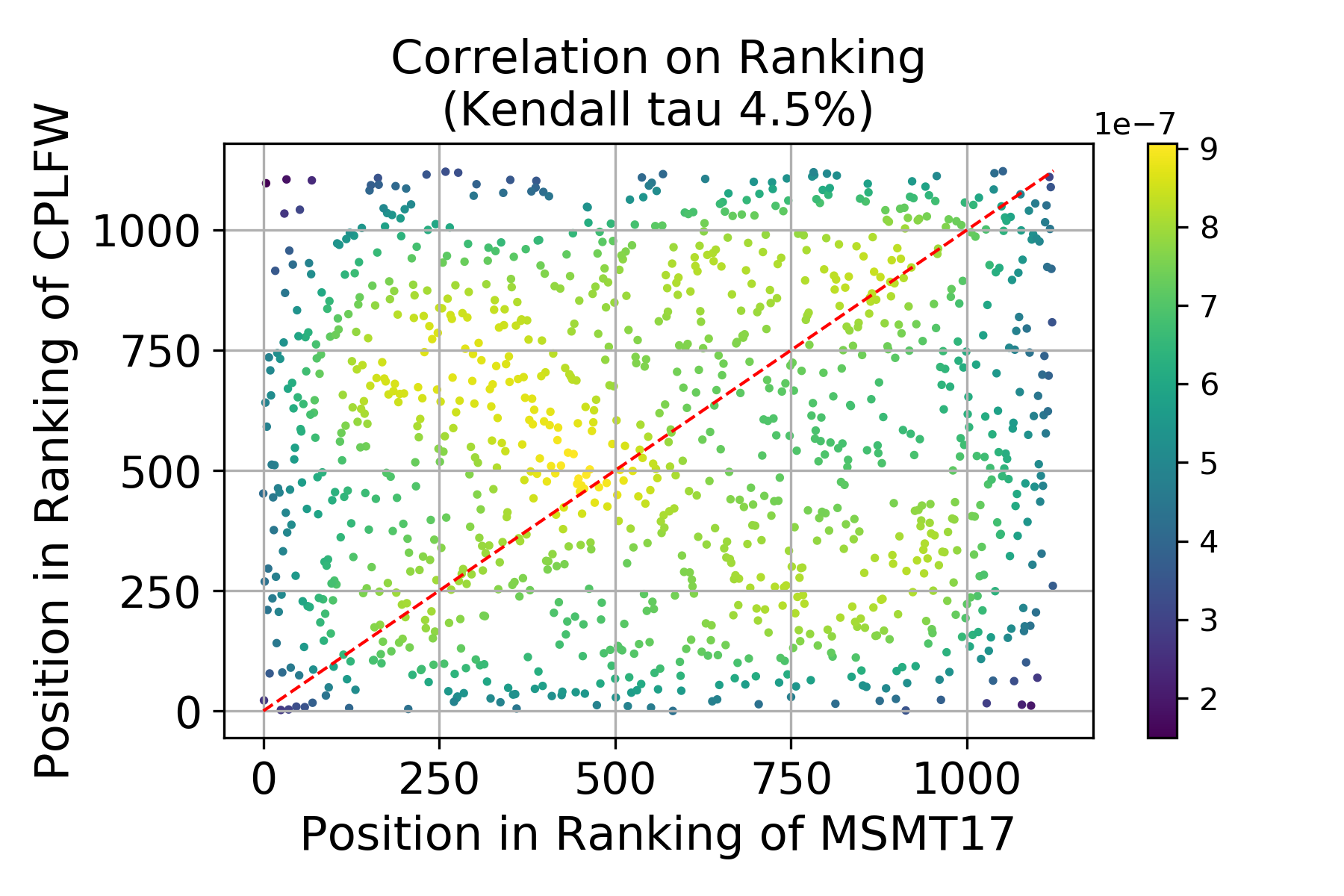}
		\end{minipage}
	}
	\subfigure[Person and Products]{
		\begin{minipage}[t]{0.48\linewidth}
			\centering
			\includegraphics[width=0.99\linewidth]{image/Rank_correlation-MSMT17-SOP-50.3.png}
		\end{minipage}
	}
	\subfigure[Vechicle and Products]{
		\begin{minipage}[t]{0.48\linewidth}
			\centering
			\includegraphics[width=0.99\linewidth]{image/Rank_correlation-VeriWild-SOP-48.7.png}
		\end{minipage}
	} \ 
	\subfigure[Person and Vehicle]{
		\begin{minipage}[t]{0.48\linewidth}
			\centering
			\includegraphics[width=0.99\linewidth]{image/Rank_correlation-MSMT17-VeriWild-62.9.png}
		\end{minipage}
	}	
	\caption{All cross tasks correlations part two.}
	\label{cross_2}
\end{figure*}

\begin{figure*}[h!] 
	\subfigure[Person and Products ]{
		\begin{minipage}[t]{0.48\linewidth}
			\centering
			\includegraphics[width=0.99\linewidth]{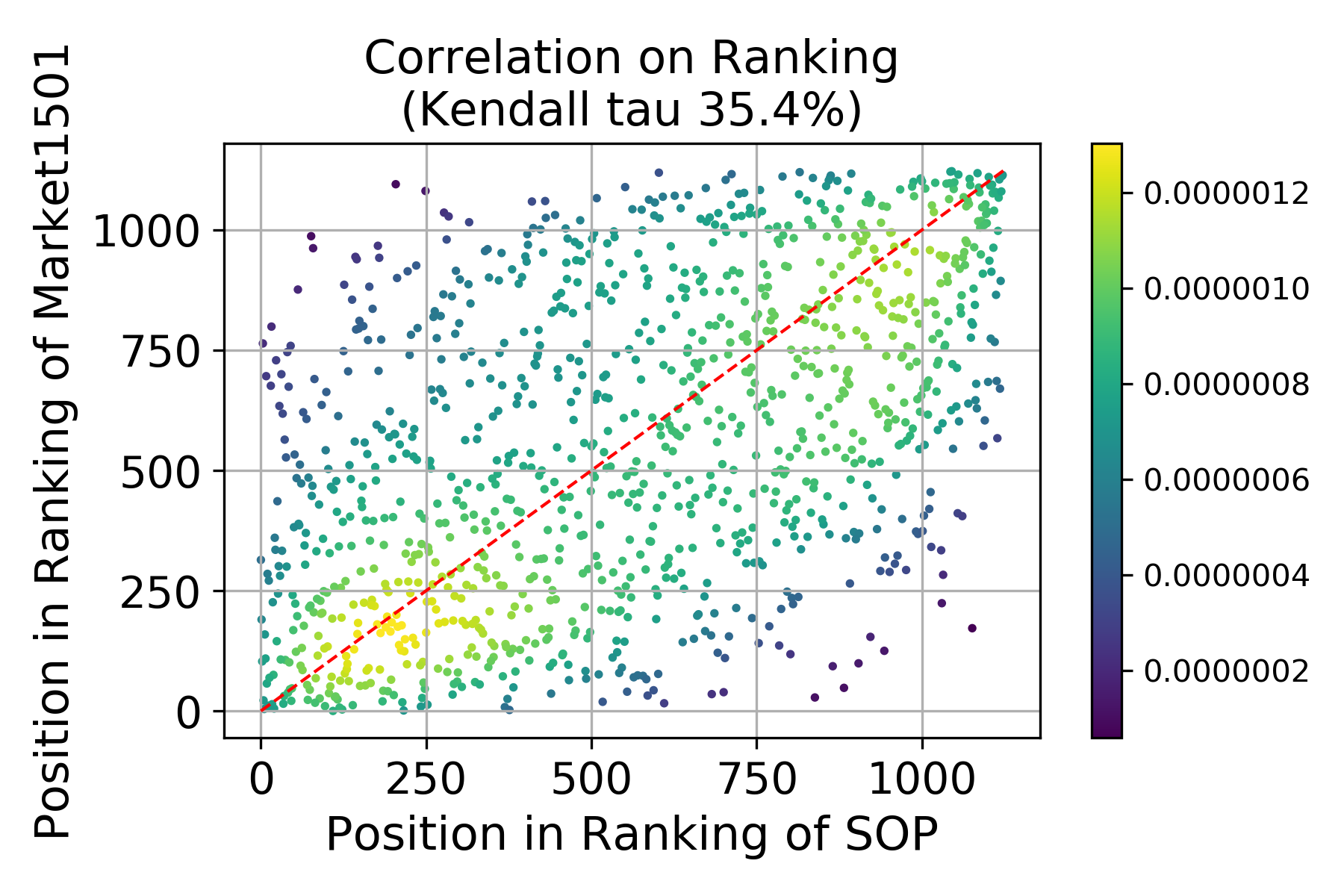}
		\end{minipage}
	}
	\subfigure[Vehicle and Products]{
		\begin{minipage}[t]{0.48\linewidth}
			\centering
			\includegraphics[width=0.99\linewidth]{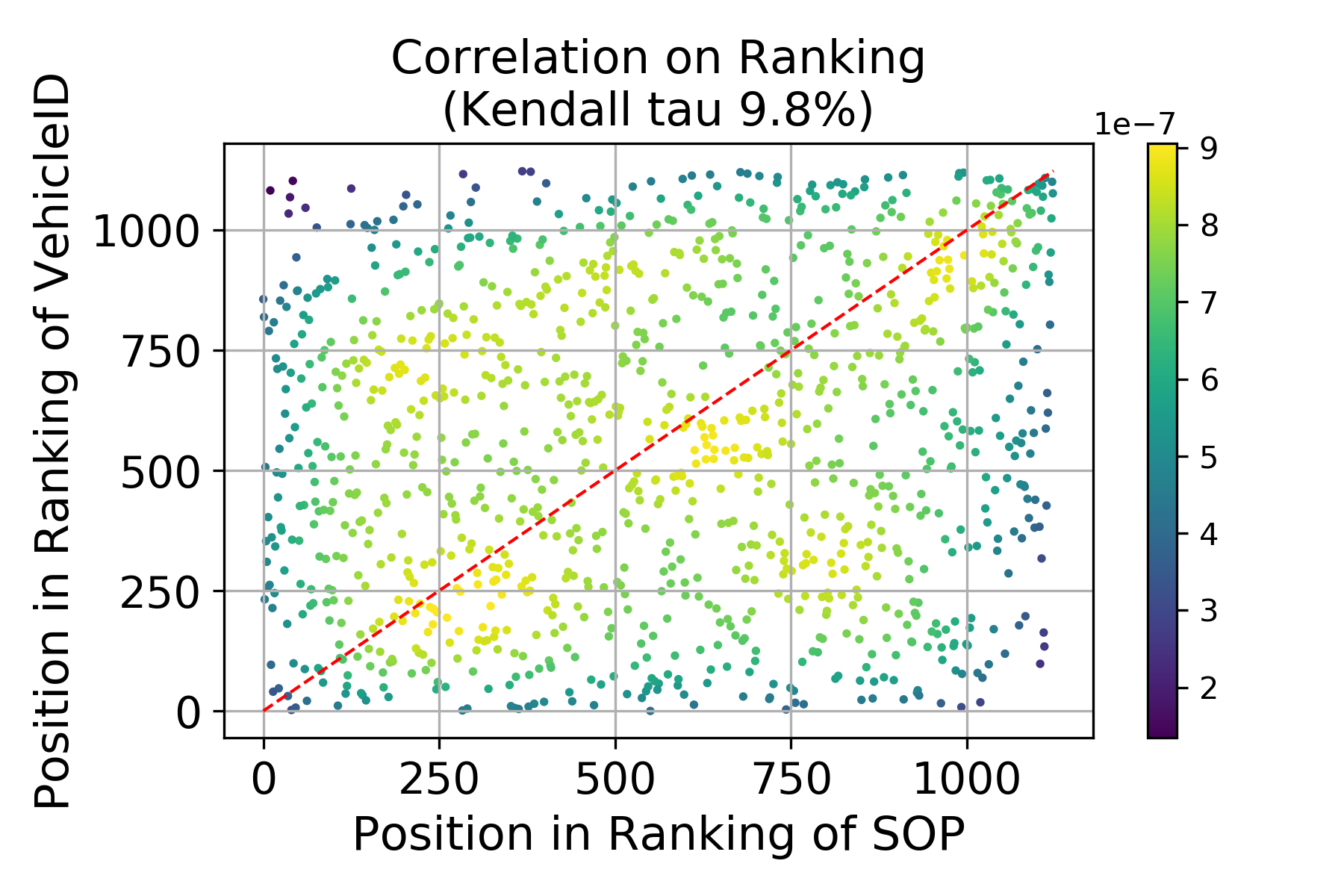}
		\end{minipage}
	}	
	\subfigure[Person and Vehicle]{
		\begin{minipage}[t]{0.48\linewidth}
			\centering
			\includegraphics[width=0.99\linewidth]{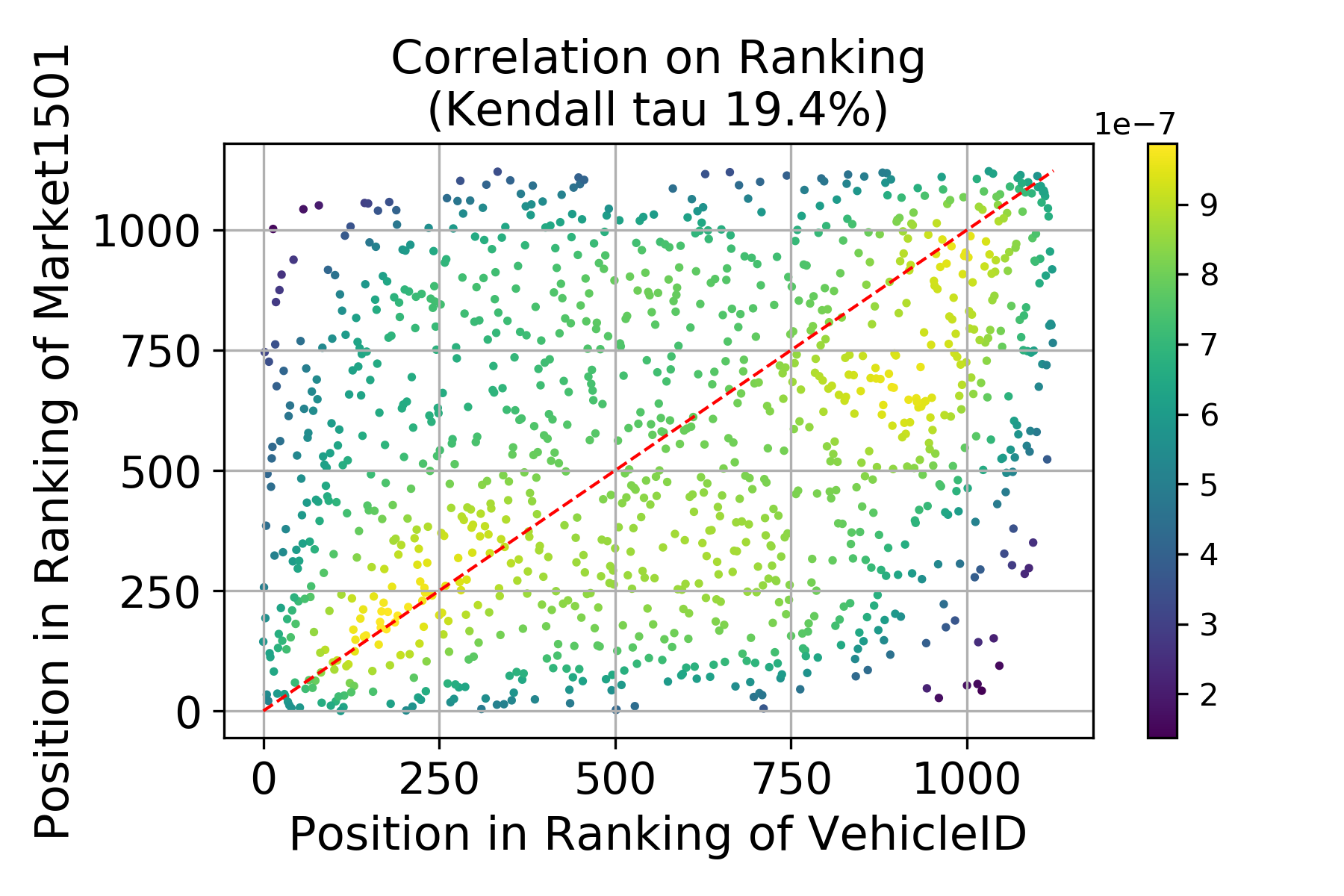}
		\end{minipage}
	}
		\subfigure[Person and Vehicle]{
		\begin{minipage}[t]{0.48\linewidth}
			\centering
			\includegraphics[width=0.99\linewidth]{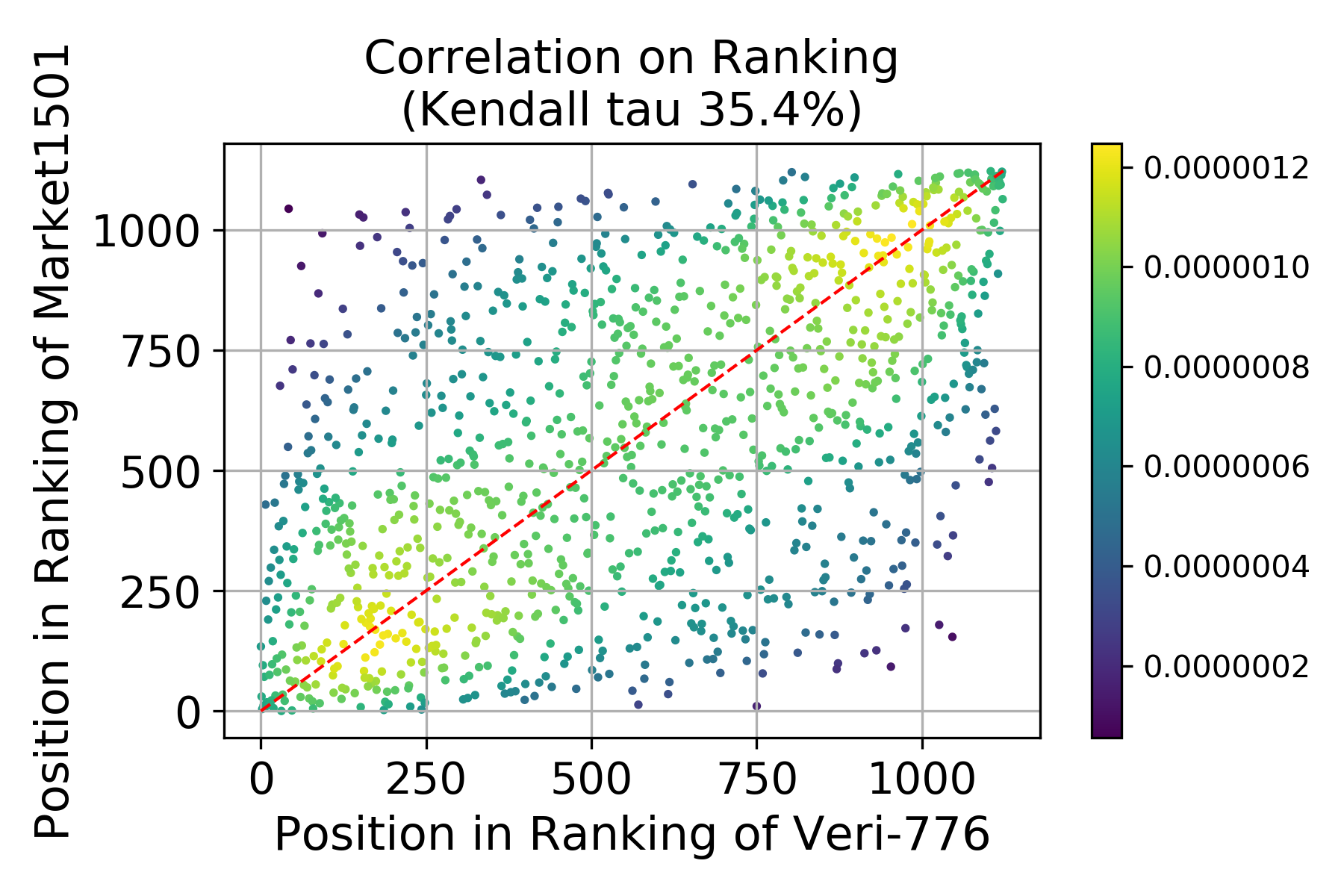}
		\end{minipage}
	}
	\subfigure[Person and Vehicle]{
		\begin{minipage}[t]{0.48\linewidth}
			\centering
			\includegraphics[width=0.99\linewidth]{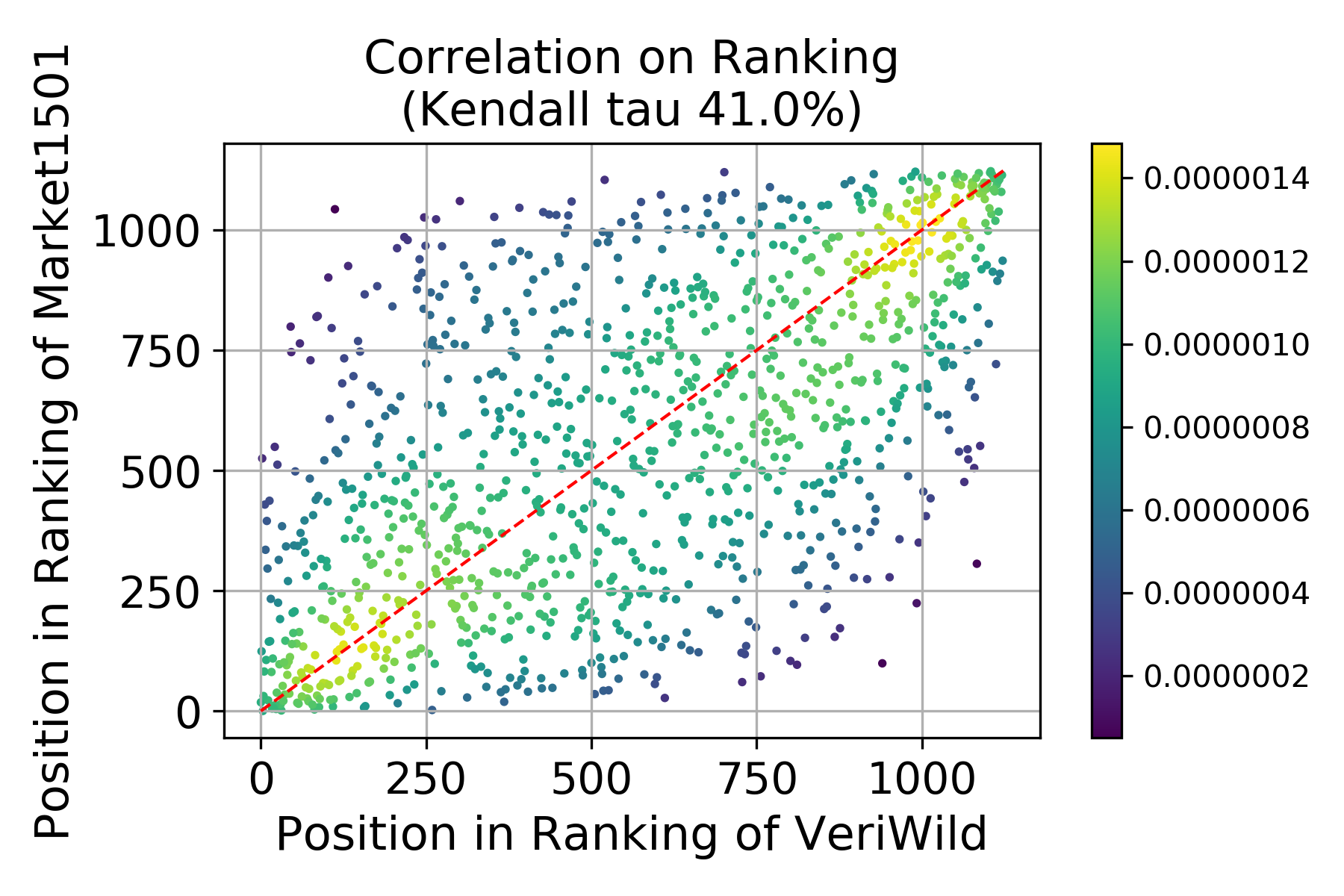}
		\end{minipage}
	}
	\subfigure[Person and Vehicle]{
		\begin{minipage}[t]{0.48\linewidth}
			\centering
			\includegraphics[width=0.99\linewidth]{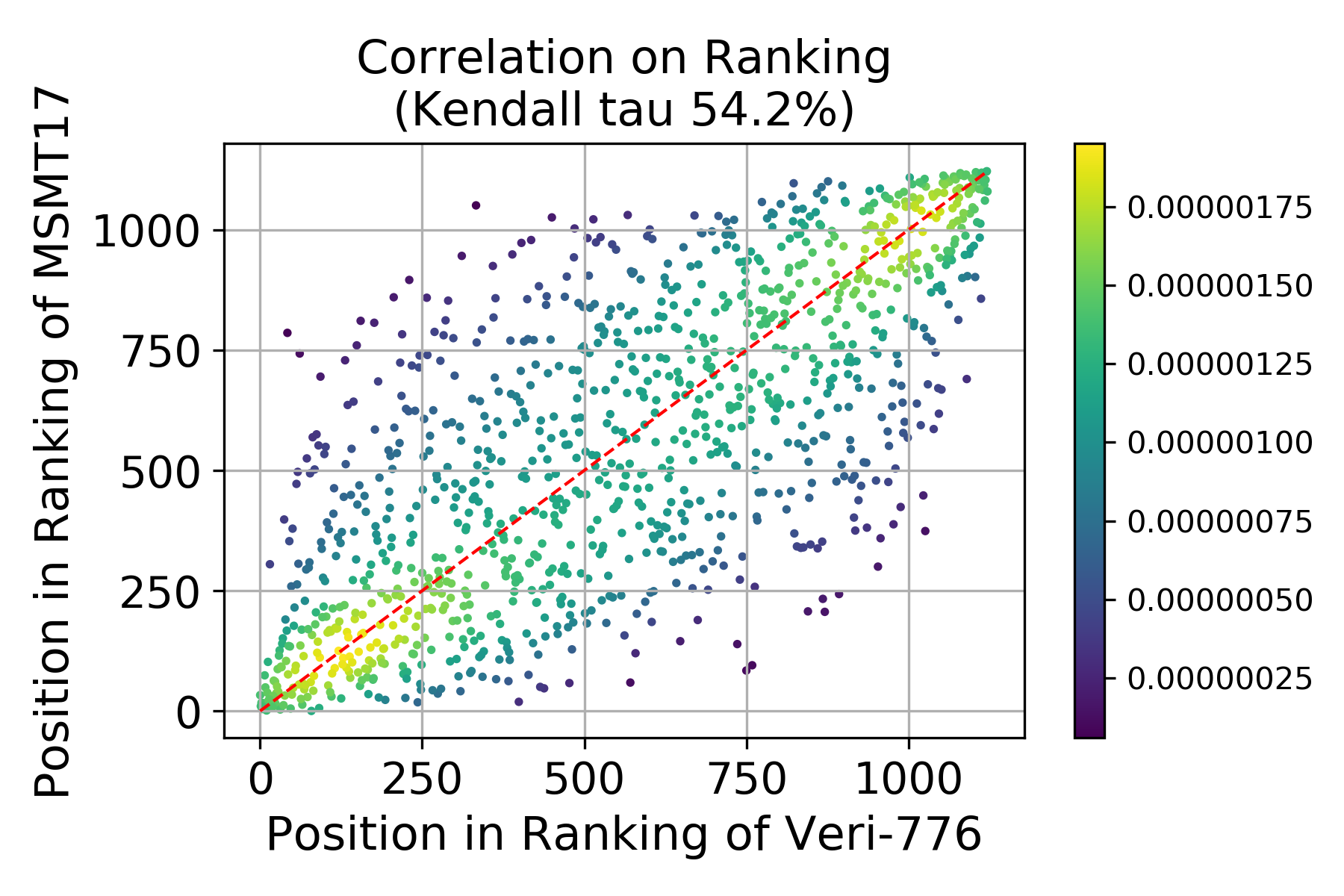}
		\end{minipage}
	}
	\subfigure[Person and Vehicle]{
		\begin{minipage}[t]{0.48\linewidth}
			\centering
			\includegraphics[width=0.99\linewidth]{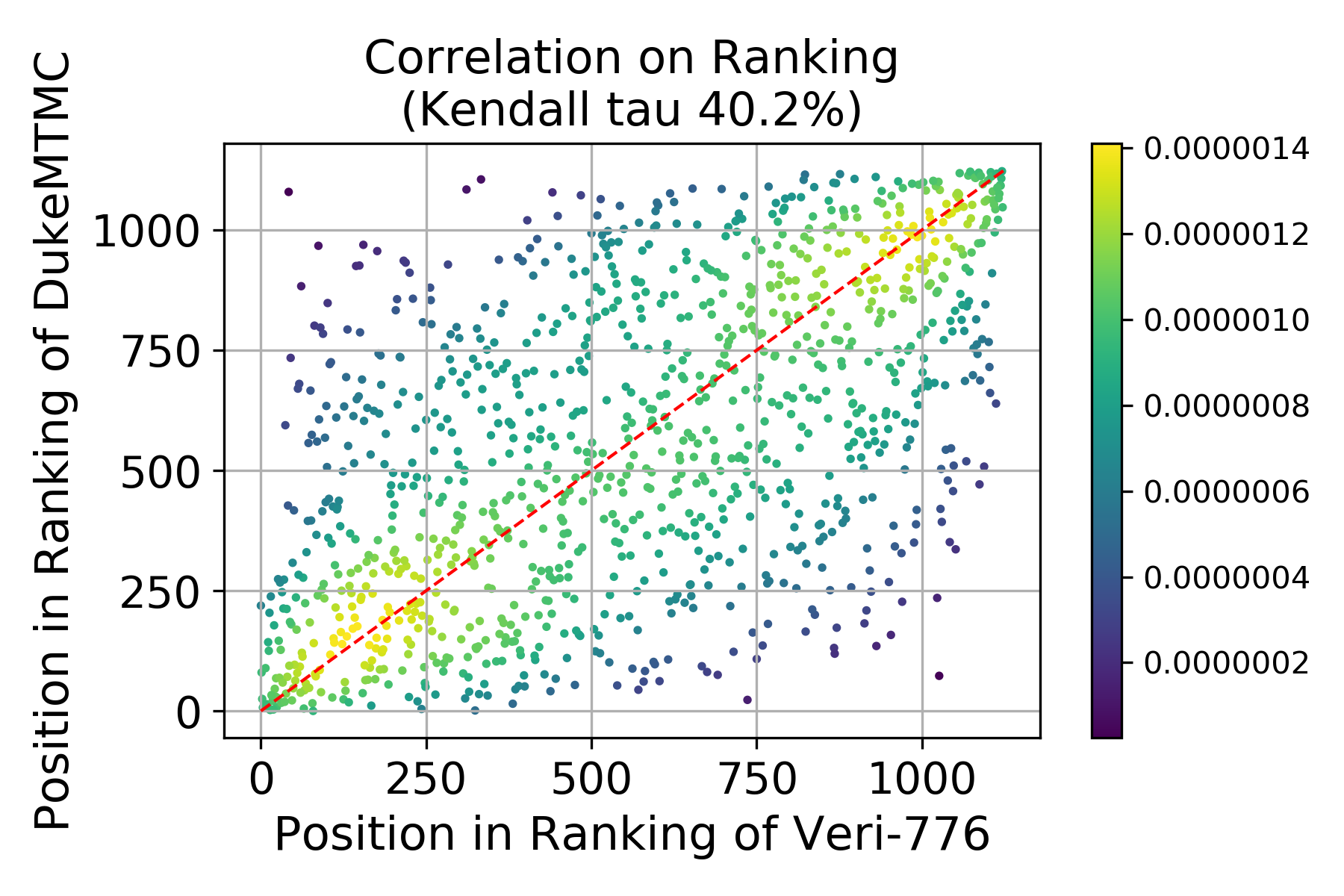}
		\end{minipage}
	} \ 
	\subfigure[Person and Vehicle]{
		\begin{minipage}[t]{0.48\linewidth}
			\centering
			\includegraphics[width=0.99\linewidth]{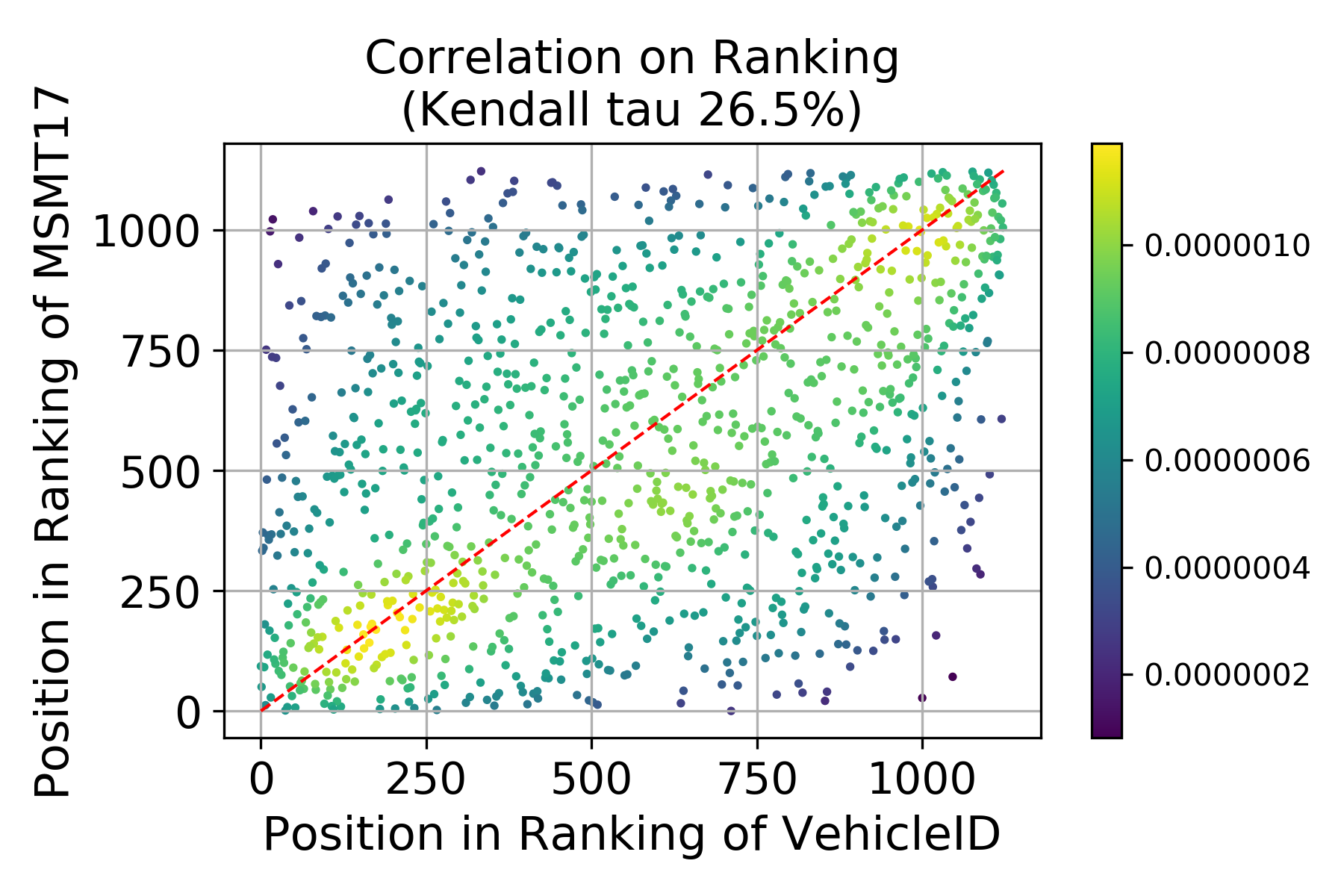}
		\end{minipage}
	}	
	\caption{All cross tasks correlations part three.}
	\label{cross_3}
\end{figure*}

\begin{figure*}[h!]
	\subfigure[Predictions on MTMC.] {
		\begin{minipage}[t]{0.48\linewidth}
			\centering
			\includegraphics[width=0.99\linewidth]{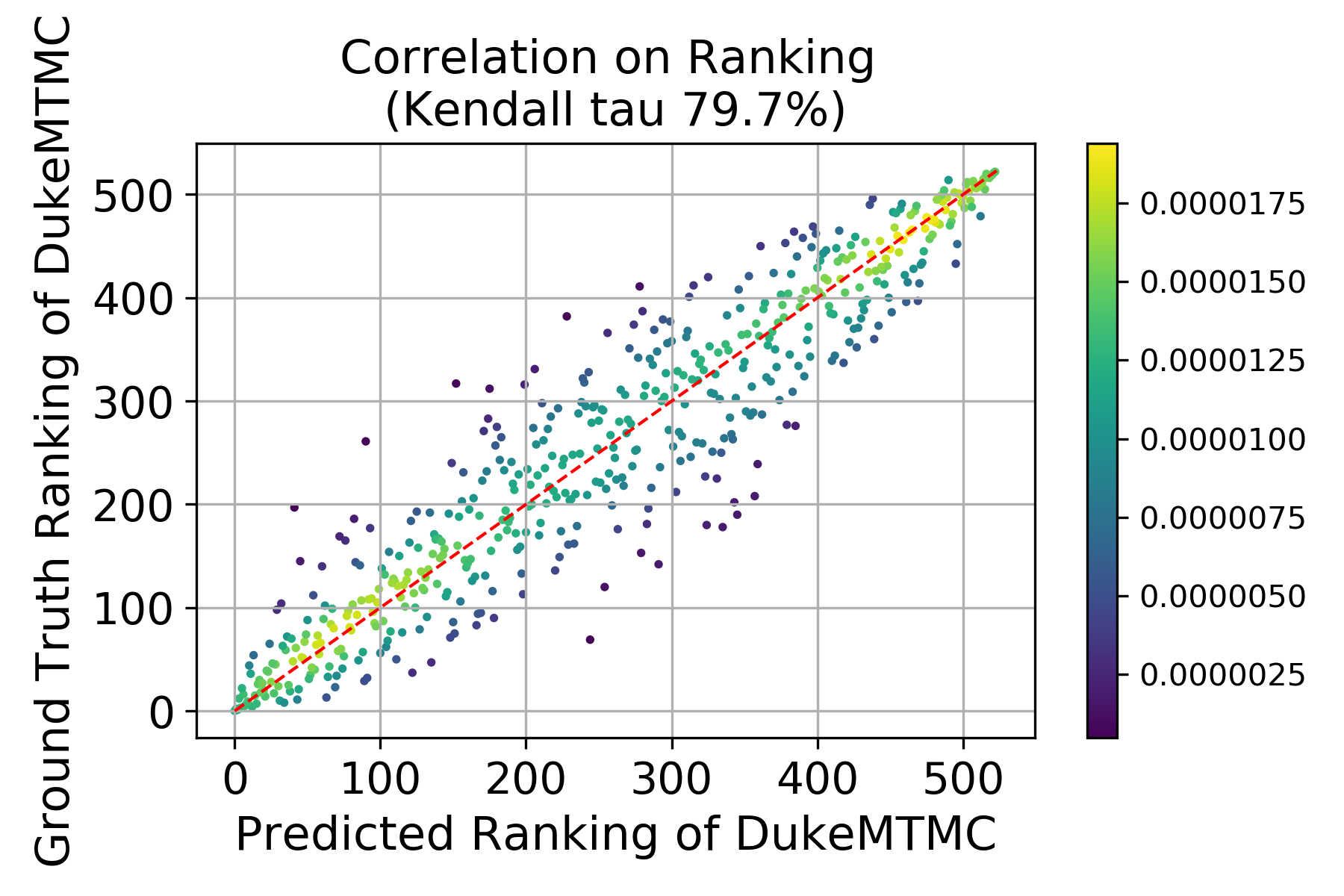}
		\end{minipage}
	}
	\subfigure[Predictions on MSMT17. ]{
		\begin{minipage}[t]{0.48\linewidth}
			\centering
			\includegraphics[width=0.99\linewidth]{image/Predict_Rank_correlation-MSMT17-86.1.png}
		\end{minipage}
	}	
	\subfigure[Predictions on SOP.]{
		\begin{minipage}[t]{0.48\linewidth}
			\centering
			\includegraphics[width=0.99\linewidth]{image/Predict_Rank_correlation-SOP-79.7.png}
		\end{minipage}
	}
	\subfigure[Predictions on Veri-776.]{
		\begin{minipage}[t]{0.48\linewidth}
			\centering
			\includegraphics[width=0.99\linewidth]{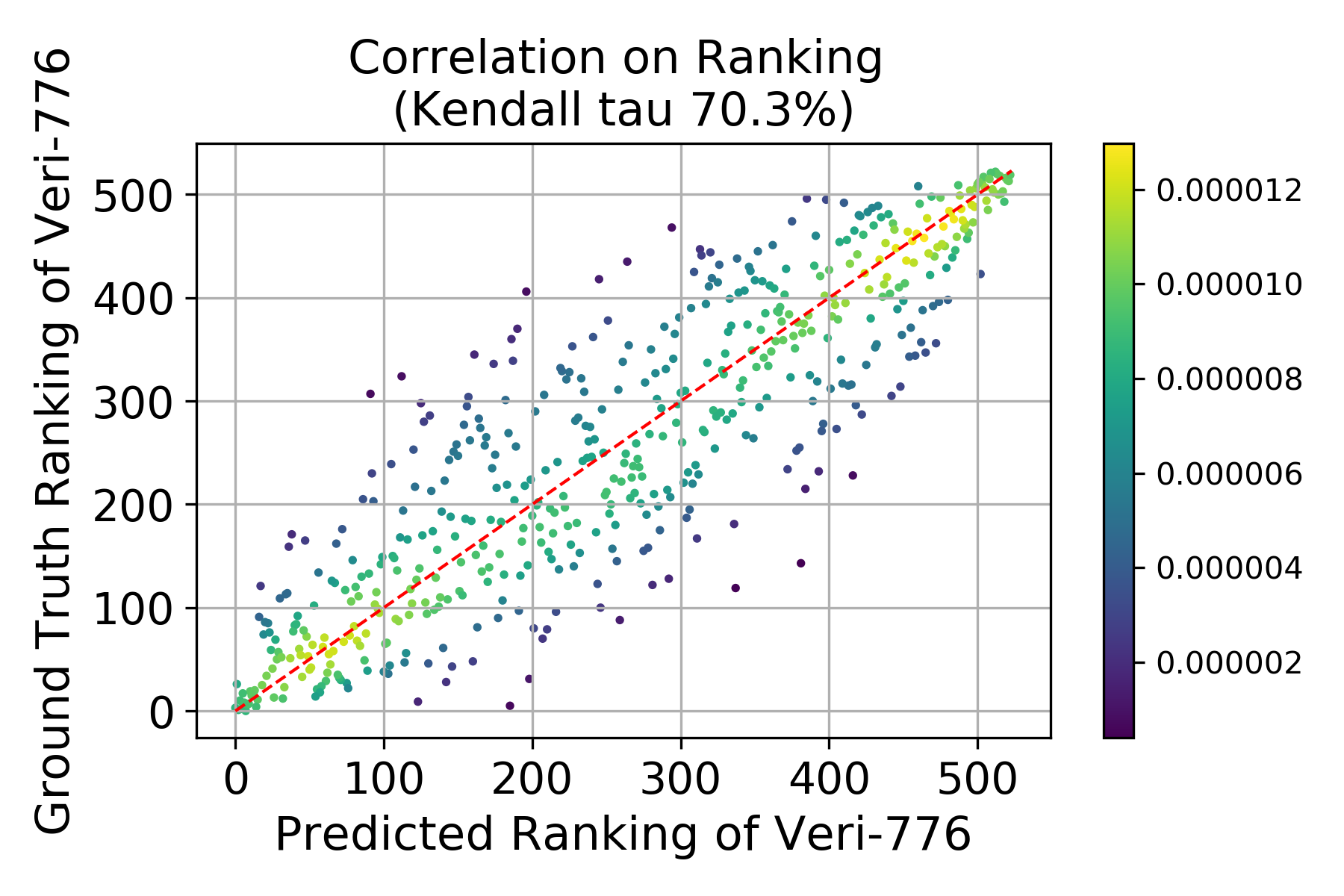}
		\end{minipage}
	}
	\subfigure[Predictions on Market1501.]{
		\begin{minipage}[t]{0.48\linewidth}
			\centering
			\includegraphics[width=0.99\linewidth]{image/Predict_Rank_correlation-Market1501-78.3.png}
		\end{minipage}
	} \ 
	\subfigure[Predictions on VeriWild.]{
		\begin{minipage}[t]{0.48\linewidth}
			\centering
			\includegraphics[width=0.99\linewidth]{image/Predict_Rank_correlation-VeriWild-74.6.png}
		\end{minipage}
	}
	
	\subfigure[Predictions on VehicleID.]{
		\begin{minipage}[t]{0.48\linewidth}
			\centering
			\includegraphics[width=0.99\linewidth]{image/Predict_Rank_correlation-Veri-776-70.3.png} 
		\end{minipage}
	}
	\subfigure[Predictions on CPLFW.]{
		\begin{minipage}[t]{0.48\linewidth}
			\centering
			\includegraphics[width=0.99\linewidth]{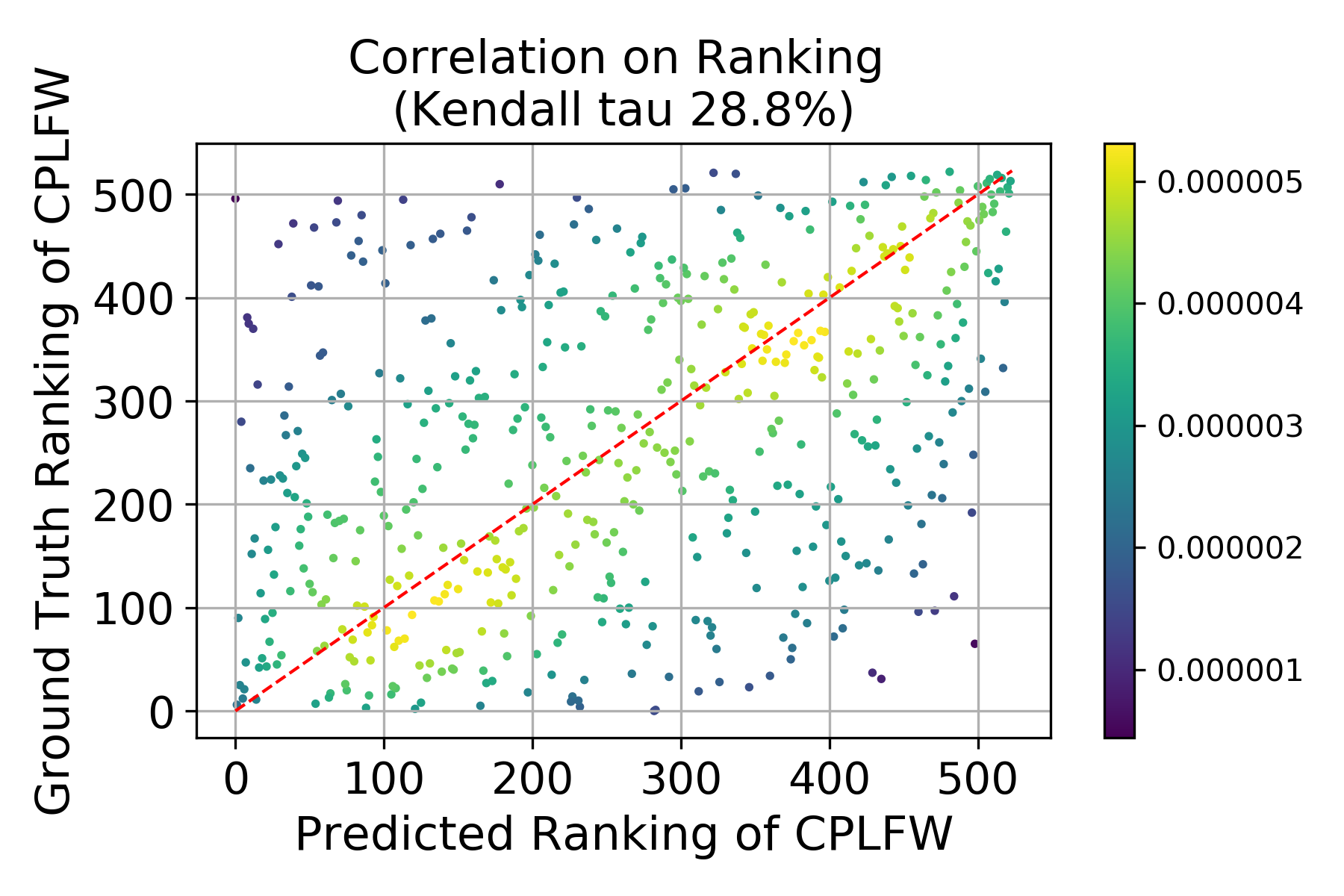}
		\end{minipage}
	}
 	\caption{Performance of task specific predictors on more benchmarks.}\label{rank_prediction_more}
\end{figure*}

\section{All cross tasks correlations}

In this section, we will show all cross tasks correlations. As shown in figure \ref{cross_1}, \ref{cross_2} and \ref{cross_3}, the benchmark of face is sightly correlated to all other tasks. While, certain benchmarks of person, vehicle and products are highly correlated. 

 \section{The performances of task specific predictors} Figure \ref{rank_prediction_more}, illustrate the performances of the task specific predictors. We measure the correlation between the predicted rankings and ground truth rankings of selected benchmarks. As shown in the figure, the predictors all have very good accuracy except for CPLFW.

\clearpage

\bibliographystyle{splncs04}
\bibliography{ref}

\begin{thebibliography}{10}
\providecommand{\url}[1]{\texttt{#1}}
\providecommand{\urlprefix}{URL }
\providecommand{\doi}[1]{https://doi.org/#1}

\bibitem{abnar2021exploring}
Abnar, S., Dehghani, M., Neyshabur, B., Sedghi, H.: Exploring the limits of
  large scale pre-training. arXiv preprint arXiv:2110.02095  (2021)

\bibitem{an2021partial}
An, X., Zhu, X., Gao, Y., Xiao, Y., Zhao, Y., Feng, Z., Wu, L., Qin, B., Zhang,
  M., Zhang, D., et~al.: Partial fc: Training 10 million identities on a single
  machine. In: Proceedings of the IEEE/CVF International Conference on Computer
  Vision. pp. 1445--1449 (2021)

\bibitem{bommasani2021opportunities}
Bommasani, R., Hudson, D.A., Adeli, E., Altman, R., Arora, S., von Arx, S.,
  Bernstein, M.S., Bohg, J., Bosselut, A., Brunskill, E., et~al.: On the
  opportunities and risks of foundation models. arXiv preprint arXiv:2108.07258
   (2021)

\bibitem{cai2019once}
Cai, H., Gan, C., Wang, T., Zhang, Z., Han, S.: Once-for-all: Train one network
  and specialize it for efficient deployment. arXiv preprint arXiv:1908.09791
  (2019)

\bibitem{chen2021autoformer}
Chen, M., Peng, H., Fu, J., Ling, H.: Autoformer: Searching transformers for
  visual recognition. In: Proceedings of the IEEE/CVF International Conference
  on Computer Vision. pp. 12270--12280 (2021)

\bibitem{chen2018gradnorm}
Chen, Z., Badrinarayanan, V., Lee, C.Y., Rabinovich, A.: Gradnorm: Gradient
  normalization for adaptive loss balancing in deep multitask networks. In:
  International Conference on Machine Learning. pp. 794--803. PMLR (2018)

\bibitem{chrysos2021deep}
Chrysos, G.G., Moschoglou, S., Bouritsas, G., Deng, J., Panagakis, Y.,
  Zafeiriou, S.P.: Deep polynomial neural networks. IEEE Transactions on
  Pattern Analysis and Machine Intelligence  (2021)

\bibitem{deng2020retinaface}
Deng, J., Guo, J., Ververas, E., Kotsia, I., Zafeiriou, S.: Retinaface:
  Single-shot multi-level face localisation in the wild. In: Proceedings of the
  IEEE/CVF Conference on Computer Vision and Pattern Recognition. pp.
  5203--5212 (2020)

\bibitem{deng2019lightweight}
Deng, J., Guo, J., Zhang, D., Deng, Y., Lu, X., Shi, S.: Lightweight face
  recognition challenge. In: Proceedings of the IEEE/CVF International
  Conference on Computer Vision Workshops. pp.~0--0 (2019)

\bibitem{dosovitskiy2020image}
Dosovitskiy, A., Beyer, L., Kolesnikov, A., Weissenborn, D., Zhai, X.,
  Unterthiner, T., Dehghani, M., Minderer, M., Heigold, G., Gelly, S., et~al.:
  An image is worth 16x16 words: Transformers for image recognition at scale.
  arXiv preprint arXiv:2010.11929  (2020)

\bibitem{duong2015low}
Duong, L., Cohn, T., Bird, S., Cook, P.: Low resource dependency parsing:
  Cross-lingual parameter sharing in a neural network parser. In: Proceedings
  of the 53rd annual meeting of the Association for Computational Linguistics
  and the 7th international joint conference on natural language processing
  (volume 2: short papers). pp. 845--850 (2015)

\bibitem{fedus2021switch}
Fedus, W., Zoph, B., Shazeer, N.: Switch transformers: Scaling to trillion
  parameter models with simple and efficient sparsity. arXiv preprint
  arXiv:2101.03961  (2021)

\bibitem{gao2020mtl}
Gao, Y., Bai, H., Jie, Z., Ma, J., Jia, K., Liu, W.: Mtl-nas: Task-agnostic
  neural architecture search towards general-purpose multi-task learning. In:
  Proceedings of the IEEE/CVF Conference on computer vision and pattern
  recognition. pp. 11543--11552 (2020)

\bibitem{guo2018dynamic}
Guo, M., Haque, A., Huang, D.A., Yeung, S., Fei-Fei, L.: Dynamic task
  prioritization for multitask learning. In: Proceedings of the European
  conference on computer vision (ECCV). pp. 270--287 (2018)

\bibitem{guo2016ms}
Guo, Y., Zhang, L., Hu, Y., He, X., Gao, J.: Ms-celeb-1m: A dataset and
  benchmark for large-scale face recognition. In: European conference on
  computer vision. pp. 87--102. Springer (2016)

\bibitem{hazimeh2021dselect}
Hazimeh, H., Zhao, Z., Chowdhery, A., Sathiamoorthy, M., Chen, Y., Mazumder,
  R., Hong, L., Chi, E.: Dselect-k: Differentiable selection in the mixture of
  experts with applications to multi-task learning. Advances in Neural
  Information Processing Systems  \textbf{34},  29335--29347 (2021)

\bibitem{he2020fastreid}
He, L., Liao, X., Liu, W., Liu, X., Cheng, P., Mei, T.: Fastreid: A pytorch
  toolbox for general instance re-identification. arXiv preprint
  arXiv:2006.02631  (2020)

\bibitem{he2021transreid}
He, S., Luo, H., Wang, P., Wang, F., Li, H., Jiang, W.: Transreid:
  Transformer-based object re-identification. In: Proceedings of the IEEE/CVF
  International Conference on Computer Vision. pp. 15013--15022 (2021)

\bibitem{herzog2021lightweight}
Herzog, F., Ji, X., Teepe, T., H{\"o}rmann, S., Gilg, J., Rigoll, G.:
  Lightweight multi-branch network for person re-identification. In: 2021 IEEE
  International Conference on Image Processing (ICIP). pp. 1129--1133. IEEE
  (2021)

\bibitem{hou2020dynabert}
Hou, L., Huang, Z., Shang, L., Jiang, X., Chen, X., Liu, Q.: Dynabert: Dynamic
  bert with adaptive width and depth. Advances in Neural Information Processing
  Systems  \textbf{33},  9782--9793 (2020)

\bibitem{huang2008labeled}
Huang, G.B., Mattar, M., Berg, T., Learned-Miller, E.: Labeled faces in the
  wild: A database forstudying face recognition in unconstrained environments.
  In: Workshop on faces in'Real-Life'Images: detection, alignment, and
  recognition (2008)

\bibitem{huynh2021strong}
Huynh, S.V.: A strong baseline for vehicle re-identification. In: Proceedings
  of the IEEE/CVF Conference on Computer Vision and Pattern Recognition. pp.
  4147--4154 (2021)

\bibitem{kendall2018multi}
Kendall, A., Gal, Y., Cipolla, R.: Multi-task learning using uncertainty to
  weigh losses for scene geometry and semantics. In: Proceedings of the IEEE
  conference on computer vision and pattern recognition. pp. 7482--7491 (2018)

\bibitem{kim2020groupface}
Kim, Y., Park, W., Roh, M.C., Shin, J.: Groupface: Learning latent groups and
  constructing group-based representations for face recognition. In:
  Proceedings of the IEEE/CVF Conference on Computer Vision and Pattern
  Recognition. pp. 5621--5630 (2020)

\bibitem{kudugunta2021beyond}
Kudugunta, S., Huang, Y., Bapna, A., Krikun, M., Lepikhin, D., Luong, M.T.,
  Firat, O.: Beyond distillation: Task-level mixture-of-experts for efficient
  inference. arXiv preprint arXiv:2110.03742  (2021)

\bibitem{lee2020compounding}
Lee, J., Won, T., Lee, T.K., Lee, H., Gu, G., Hong, K.: Compounding the
  performance improvements of assembled techniques in a convolutional neural
  network. arXiv preprint arXiv:2001.06268  (2020)

\bibitem{li2020gp}
Li, Z., Xi, T., Deng, J., Zhang, G., Wen, S., He, R.: Gp-nas: Gaussian process
  based neural architecture search. In: Proceedings of the IEEE/CVF Conference
  on Computer Vision and Pattern Recognition. pp. 11933--11942 (2020)

\bibitem{liu2016deep}
Liu, H., Tian, Y., Yang, Y., Pang, L., Huang, T.: Deep relative distance
  learning: Tell the difference between similar vehicles. In: Proceedings of
  the IEEE conference on computer vision and pattern recognition. pp.
  2167--2175 (2016)

\bibitem{liu2016recurrent}
Liu, P., Qiu, X., Huang, X.: Recurrent neural network for text classification
  with multi-task learning. arXiv preprint arXiv:1605.05101  (2016)

\bibitem{liu2019multi}
Liu, X., He, P., Chen, W., Gao, J.: Multi-task deep neural networks for natural
  language understanding. arXiv preprint arXiv:1901.11504  (2019)

\bibitem{long2015learning}
Long, M., Wang, J.: Learning multiple tasks with deep relationship networks.
  arXiv preprint arXiv:1506.02117  \textbf{2}(1) (2015)

\bibitem{lou2019veri}
Lou, Y., Bai, Y., Liu, J., Wang, S., Duan, L.: Veri-wild: A large dataset and a
  new method for vehicle re-identification in the wild. In: Proceedings of the
  IEEE/CVF conference on computer vision and pattern recognition. pp.
  3235--3243 (2019)

\bibitem{lu2017fully}
Lu, Y., Kumar, A., Zhai, S., Cheng, Y., Javidi, T., Feris, R.: Fully-adaptive
  feature sharing in multi-task networks with applications in person attribute
  classification. In: Proceedings of the IEEE conference on computer vision and
  pattern recognition. pp. 5334--5343 (2017)

\bibitem{misra2016cross}
Misra, I., Shrivastava, A., Gupta, A., Hebert, M.: Cross-stitch networks for
  multi-task learning. In: Proceedings of the IEEE conference on computer
  vision and pattern recognition. pp. 3994--4003 (2016)

\bibitem{moschoglou2017agedb}
Moschoglou, S., Papaioannou, A., Sagonas, C., Deng, J., Kotsia, I., Zafeiriou,
  S.: Agedb: the first manually collected, in-the-wild age database. In:
  proceedings of the IEEE conference on computer vision and pattern recognition
  workshops. pp. 51--59 (2017)

\bibitem{oh2016deep}
Oh~Song, H., Xiang, Y., Jegelka, S., Savarese, S.: Deep metric learning via
  lifted structured feature embedding. In: Proceedings of the IEEE conference
  on computer vision and pattern recognition. pp. 4004--4012 (2016)

\bibitem{sener2018multi}
Sener, O., Koltun, V.: Multi-task learning as multi-objective optimization.
  Advances in neural information processing systems  \textbf{31} (2018)

\bibitem{sengupta2016frontal}
Sengupta, S., Chen, J.C., Castillo, C., Patel, V.M., Chellappa, R., Jacobs,
  D.W.: Frontal to profile face verification in the wild. In: 2016 IEEE winter
  conference on applications of computer vision (WACV). pp.~1--9. IEEE (2016)

\bibitem{subramanian2018learning}
Subramanian, S., Trischler, A., Bengio, Y., Pal, C.J.: Learning general purpose
  distributed sentence representations via large scale multi-task learning.
  arXiv preprint arXiv:1804.00079  (2018)

\bibitem{suteu2019regularizing}
Suteu, M., Guo, Y.: Regularizing deep multi-task networks using orthogonal
  gradients. arXiv preprint arXiv:1912.06844  (2019)

\bibitem{wang2020hat}
Wang, H., Wu, Z., Liu, Z., Cai, H., Zhu, L., Gan, C., Han, S.: Hat:
  Hardware-aware transformers for efficient natural language processing. arXiv
  preprint arXiv:2005.14187  (2020)

\bibitem{wei2018person}
Wei, L., Zhang, S., Gao, W., Tian, Q.: Person transfer gan to bridge domain gap
  for person re-identification. In: Proceedings of the IEEE conference on
  computer vision and pattern recognition. pp. 79--88 (2018)

\bibitem{yu2020gradient}
Yu, T., Kumar, S., Gupta, A., Levine, S., Hausman, K., Finn, C.: Gradient
  surgery for multi-task learning. Advances in Neural Information Processing
  Systems  \textbf{33},  5824--5836 (2020)

\bibitem{zheng2015person}
Zheng, L., Shen, L., Tian, L., Wang, S., Bu, J., Tian, Q.: Person
  re-identification meets image search. arXiv preprint arXiv:1502.02171  (2015)

\bibitem{zheng2018cross}
Zheng, T., Deng, W.: Cross-pose lfw: A database for studying cross-pose face
  recognition in unconstrained environments. Beijing University of Posts and
  Telecommunications, Tech. Rep  \textbf{5}, ~7 (2018)

\bibitem{zheng2017cross}
Zheng, T., Deng, W., Hu, J.: Cross-age lfw: A database for studying cross-age
  face recognition in unconstrained environments. arXiv preprint
  arXiv:1708.08197  (2017)

\end{thebibliography}
\end{document}